\documentclass[preprint,12pt]{elsarticle}

\usepackage[a4paper, margin=0.8in]{geometry}

\usepackage{amssymb}

\usepackage{booktabs}                   
\usepackage{tabularx}                   
\usepackage{longtable}                  
\usepackage{ltxtable}                   
\usepackage[linesnumbered,ruled,vlined]{algorithm2e}

\usepackage{algorithmic}                

\usepackage{relsize}                    

\usepackage{pgfplots}

\usepackage{graphicx}                   

\usepackage{subfig}                     
\usepackage{amsmath}  
\usepackage{amssymb}  

\newif\ifcomm

\commtrue 
\ifcomm
\else
\commfalse
\fi
\ifcomm
\newcommand\ak[1]{\textcolor{orange}{AK: #1}}
\else
\newcommand\ak[1]{}
\fi

\journal{Journal Name}

\begin{document}

\pgfplotsset{compat=1.18}
\usepgfplotslibrary{fillbetween}

\begin{frontmatter}



\title{Smart Manufacturing: MLOps-Enabled Event-Driven Architecture for Enhanced Control in Steel Production}


\author[inst1]{Bestoun S. Ahmed}

\affiliation[inst1]{Karlstad University, Computer Science Department,
            city={Karlstad},
            postcode={65635}, 
            country={Sweden}}

\author[inst2]{Tommaso Azzalin}

\affiliation[inst2]{University of Bologna, Department of Computer Science and Engineering,
            city={Bologna},
            postcode={40126}, 
            country={Italy}}

\author[inst1,inst3]{Andreas Kassler}

\affiliation[inst3]{Deggendorf Institute of Technology, Faculty of Applied Computer Science,
            city={Deggendorf},
            postcode={94469}, 
            country={Germany}}

\author[inst4]{Andreas Thore}
\affiliation[inst4]{Smart Industrial Automation, RISE Research Institutes of Sweden,
            city={Västerås},
            country={Sweden}}

\author[inst5]{Hans Lindbäck}
\affiliation[inst5]{Bharat Forge Kilsta AB,
            city={Karlskoga},
            country={Sweden}}

\begin{abstract}

We explore a Digital Twin-Based Approach for Smart Manufacturing to improve Sustainability, Efficiency, and Cost-Effectiveness for a steel production plant. Our system is based on a micro-service edge-compute platform that ingests real-time sensor data from the process into a digital twin over a converged network infrastructure. We implement agile machine learning-based control loops in the digital twin to optimize induction furnace heating, enhance operational quality, and reduce process waste. Key to our approach is a Deep Reinforcement learning-based agent used in our machine learning operation (MLOps) driven system to autonomously correlate the system state with its digital twin to identify correction actions that aim to optimize power settings for the plant. We present the theoretical basis, architectural details, and practical implications of our approach to reduce manufacturing waste and increase production quality. We design the system for flexibility so that our scalable event-driven architecture can be adapted to various industrial applications. With this research, we propose a pivotal step towards the transformation of traditional processes into intelligent systems, aligning with sustainability goals and emphasizing the role of MLOps in shaping the future of data-driven manufacturing.

\end{abstract}




\end{frontmatter}

\newcommand{\sectionlink}[1]{\hyperref[#1]{Section} \ref{#1}}
\newcommand{\figurelink}[1]{\hyperref[#1]{Figure} \ref{#1}}
\newcommand{\tablelink}[1]{\hyperref[#1]{Table} \ref{#1}}


\section{Introduction}\label{sec:introduction}

During the past two decades, many companies have faced production problems that have led to low efficiency, poor product quality, and high costs. In particular, traditional and heavy industries, including steel and metal processing, also face material waste challenges that have a significant environmental impact and reduce their profits. An essential reason for product defects and high cost is the low degree of automation and digitalization of existing equipment used on the shop floor \cite{Cagle2020Digitalization}. For example, in a forging shop, some of the main processes, such as heating steel bars, still depend on manual adjustment by operators who have been trained for several years. Here, poor process control can lead to material degradation due to production temperature falling outside the specification, and product quality cannot be automatically detected during production. 

Although recent advances towards the fourth industrial revolution (Industry 4.0) \cite{Qin2016industry4.0} have prompted many companies to improve productivity and increase the reliability of equipment and production lines through a greater degree of automation and digitalization, significant challenges remain to implement these technologies effectively in complex industrial environments. This paper addresses the specific scientific challenge of developing a robust and adaptive control system for steel forging processes that can operate in real-time under dynamic conditions, a problem that existing approaches have not fully resolved.

\textcolor{blue}{Traditional approaches to process control in steel manufacturing have typically relied on conventional feedback control systems, which often struggle with the complex, non-linear dynamics of industrial heating processes \cite{Ferreiro2016, Wang2018}. Recent studies have highlighted the limitations of these conventional methods, particularly their inability to adapt to varying production conditions and material properties \cite{Wang2011, Kwak2011, Chai2017}.}

As one of the emerging technologies, digital twins (DT) are becoming increasingly important in improving digitalization in the process industry due to rapidly evolving simulation and modeling capabilities \cite{Matteo2022DT}, also leveraging the vast amount of compute processing from edge/cloud infrastructures. A DT is a high-fidelity virtual model aimed at emulating physical processes and behaviors with the ability to evaluate, optimize and predict \cite{Graessler2017DT}. It also provides a real-time link between the physical and digital worlds, allowing companies to have a complete digital footprint of their products and production processes throughout the production cycle \cite{Khan2018DT}. In turn, companies can realize significant value in areas of improved operations, reduced defects, increased efficiency, and enabled predictive maintenance \cite{Panagou2022PM}. With the acquisition of real-time data, a DT can help operators understand the actual production process and make preventive decisions when an anomaly occurs \cite{LIU2021DT}. However, the integration of DTs with advanced control algorithms in industrial settings presents significant scientific and technical challenges, particularly in terms of real-time data processing, model accuracy, and system adaptability.

While intelligent algorithms can be used together with DTs to control real systems, such as optimizing energy consumption, reducing waste, and improving product quality, their effective implementation in steel production environments remains an open problem. Traditional control algorithms in steel production typically involve control theory, mathematical modeling, and statistical analysis \cite{pimenov2022resource, pata2023enhancing}. Although these methods have yielded notable improvements, they often struggle to adapt to the dynamic and complex nature of industrial environments. Recently, data-driven algorithms using Machine Learning (ML) have demonstrated their potential to solve complex dynamic control problems in various domains, from robotics to autonomous vehicles \cite{wang2022control, elallid2022comprehensive}. However, their application in optimizing industrial processes, particularly in steel production, remains largely underexplored and presents unique scientific challenges. These challenges include: (1) the formulation of appropriate data-driven algorithms that can handle the high-dimensional, non-linear dynamics of steel production processes; (2) the development of robust learning frameworks that can operate reliably in noisy, industrial environments; and (3) the integration of these advanced algorithms with existing industrial control systems.

Furthermore, the operationalization of these algorithms poses substantial challenges due to complex system architecture requirements, necessitating a detailed design process. The deployment of such algorithms requires careful attention and involves the resolution of various challenges, thereby requiring the consideration of multiple architectural decisions from the perspective of Machine Learning Operations (MLOps). It is worth mentioning that the adoption of standard, commercialized cloud solutions from providers is rendered challenging in applications like this due to security and the high latency involved. As we aim to control industrial processes, sophisticated edge-compute architecture is required to minimize the control latency. This requires developing and implementing a custom architecture tailored to address the unique demands of controlling the production process.

In this paper, we propose a novel technical solution that addresses these scientific challenges by embedding a DT of the production process into an edge compute infrastructure to optimize the control of a steel forging heating furnace. Our solution leverages a micro-services-based architecture for integrating MLOps pipelines that receive and process sensorial data from PLCs. The core scientific contribution lies in the development of a novel Deep Reinforcement Learning (DRL) framework that can learn and adapt to the complex dynamics of the steel forging process in real-time, a problem that existing approaches have not effectively solved. To this end, our work makes the following contributions:

\begin{itemize}

    \item Advanced Integration of DRL and Industry 4.0 Principles: We present a novel framework that seamlessly combines advanced DRL techniques with Industry 4.0 principles, addressing the scientific challenge of creating adaptive, real-time control systems for complex industrial processes.

    \item Innovative MLOps-Driven Architecture: We develop a unique MLOps-driven architecture that overcomes the challenges of implementing and maintaining ML models in industrial settings, providing a scientifically grounded approach to ensuring robustness and efficiency in production environments.

    \item Novel DRL Approach for Induction Furnace Heating: We introduce a new DRL algorithm specifically designed to control electrical power during the heating process of the induction furnace, addressing the scientific challenge of optimizing non-linear, high-dimensional industrial processes in real-time.

    \item Comprehensive Empirical Validation: We provide extensive empirical results that validate the scientific and practical benefits of our system, demonstrating \ak{the potential for} significant waste reduction and operational quality improvements in a real industrial setting.

\end{itemize}

The remainder of this paper is organized as follows. Section \ref{RelatedWork} details the background of the furnace and the operational process, illustrates the problem statement, and provides a comprehensive review of related work. Section \ref{MLOPsDrivenSystemSEction3} presents the MLOps-driven system architecture for digital twin-based control, outlining the key aspects of our approach. Section \ref{DRLpowedControl} delves into the DRL framework for digital twin-based process control, explaining the necessary background and implementation details. Section \ref{Deployment} describes the deployment process, including model maintenance and management updates crucial in MLOps. Section \ref{Evaluation} presents our experimental evaluation, demonstrating the effectiveness of our approach through comprehensive empirical results. Section \ref{ImpactOnSustain} discusses the impact of our system on sustainability, efficiency, and cost-effectiveness, including a detailed use case scenario. Finally, Section \ref{Conclusion} concludes the paper and outlines directions for future work.

\section{Background and Related Work}\label{RelatedWork}
In this section, we briefly introduce the necessary background on the furnace and the operational process that we aim to control from the architecture that we are proposing. We detail the problems that arise when semi-manually controlling the furnace operation. We also summarize related work and how our work relates to it. 

\subsection{Induction heating for the forging shop}\label{descriptionOfFurnaceprocess}

This research was conducted in collaboration with Bharat Forge Kilsta AB, a leading manufacturer in the forging industry located in Karlskoga, Sweden. Bharat Forge Kilsta specializes in producing high-quality forged and machined components for the heavy vehicle and industrial sectors. The company's production facility includes state-of-the-art forging equipment, including the induction heating furnace, which is the focus of this study.

\begin{figure}
    \centering
    \includegraphics[scale=0.9]{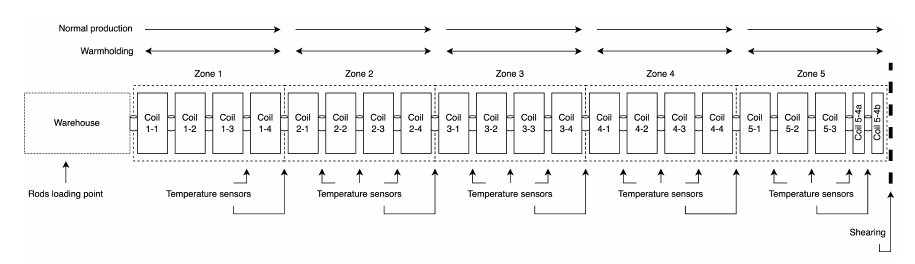}
    \caption{A graphical and simplified representation of the furnace.}
    \label{fig:RepresentationOfTheFurnce}
\end{figure}

In our work, we aim to automate the control of a furnace that is used to heat steel in a forging shop in a production plant. Steel in the plant comes in the form of rods of different lengths, diameters, and physical properties. The furnace heating these rods is built as a production line of induction coils that are approximately 1 m long and separated by a gap of about 25 cm. The induction coils are grouped into five zones of four coils each. This means that the furnace is approximately 25 m long. Figure \ref{fig:RepresentationOfTheFurnce} shows a simplified graphical representation of the furnace and its zones.

As shown in Figure \ref{fig:RepresentationOfTheFurnce}, rods that need to be inserted inside the furnace are first loaded into a warehouse located before the starting point of the production line and its dimensions are not included in the 25 m range. Roller conveyors between induction coils allow rods to move back and forth for heating. The fifth zone has five induction coils. The last induction coil is split into two halves. This splitting allows an extra roller to be placed between the coils to speed up the rod before reaching the end of the furnace, where it is sheared and produces billets of the desired length that can then go to the forging part of the plant.

The induction coils are heated through electrical power that human operators can control. The heat is induced and created into the steel by induction from the coils. \textcolor{blue}{All five zones have different powers to allow the rod to reach the appropriate target temperature when approaching the end of each of them}. Figure \ref{fig:TempratureRange} shows the minimum, maximum, and target temperatures of all five zones. The current specifications require the rod to be heated intensively in the first two zones, with even a major intensity in the second one, and then progressively reduce the power in the last three zones to reach the desired temperature on the steel rods without too many variations. To check what is happening in the furnace and to act accordingly, several sensors (pyrometers) measure the temperature that is placed at every gap between the induction coils and after shearing, some to obtain the rod positions and the velocity at which they are moving, and finally measure the power which is applied to the induction coils. Pyrometer sensors are placed to obtain quick feedback from the production line and allow human operators to act based on the retrieved values. The furnace can work in two modes that are designed for different situations:

\begin{figure}
    \centering
    \includegraphics[scale=0.9]{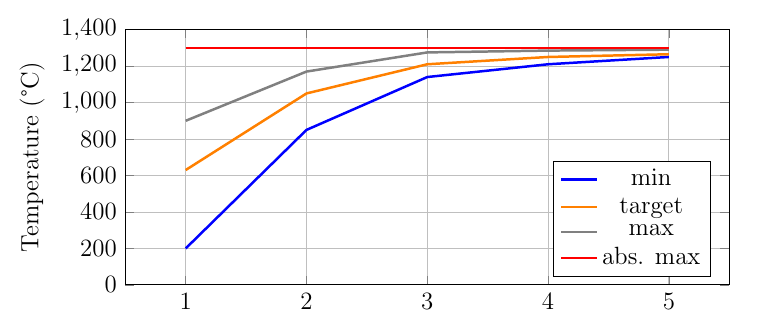}
    \caption{Temperature ranges in the furnace.}
    \label{fig:TempratureRange}
\end{figure}

\begin{itemize}
    \item \textbf{Normal production}, in which the rods are loaded from the warehouse inside the furnace and moved forward with constant speed until they reach a roller near the end of the furnace. At this point, the speed is increased and the bar is moved out of the furnace and to the shear. The electrical power applied to the induction coils is very high and follows recipes designed by the company on the basis of the diameter and material of which the rods are made. The recipes are known to work well in most cases and guide the PLC in setting the proper power for the coils in the different zones. Human operators can apply small manual adjustments to power settings based on their long experience. This straightforward mode of operation is easily manageable by human operators.

    \item \textbf{Warmholding}, in which no new rods are loaded from the warehouse. Instead, the rods inside the furnace move back and forth and are heated to a much lower temperature to avoid overheating. The goal is to allow them to go through shearing and later forging at the right temperature. There are different reasons for switching to warmholding, including furnace problems that require manual intervention. The problem is that the induction coils are separated by gaps in which the rods cool and lose heat. Therefore, different rods and parts of the same rod may have different temperatures. While back-and-forth can preserve the rods from underheating, it produces another problem called zebra temperature patterns, in which a rod has both parts at a high and a low temperature. If these patterns are too difficult to handle, operators could decide to stop the process, unloading the rods, wait for them to cool down, and start over, which would waste a lot of time and energy.    
\end{itemize}

Scrapping occurs due to large variations in the steel rods' forging temperature, mainly when the industrial plant is restarted after downtime or during warmholding mode. The rods inside the furnace that are already partially heated need to maintain their temperature without being damaged or losing heat, as this would mean a great loss of time and energy. The forging temperature window is very narrow, and having material that is under- or overheated leads to poor-quality products. Human operators control the forging process and can easily keep the steel within the temperature window during the normal production scenario, but struggle when the rod should be kept in the forging system for a certain time with a limited temperature range during maintenance and warmholding mode.

\subsection{Related Work}

Efforts to improve the efficiency, sustainability, and cost effectiveness of steel production have been the subject of extensive research. Optimizing steel production processes has long been challenging due to its significant impact on industrial sectors and the environment. The complexity of steel manufacturing, which involves various stages of intensive energy, requires continuous innovation to improve operational efficiency while minimizing environmental impact. He and Wang \cite{HE20171022} provided a comprehensive review of current practices and challenges in optimizing energy efficiency within the steel industry, identifying key areas such as energy recovery, process optimization, and waste minimization. Their work underscores the importance of energy optimization in reducing operational costs and as a critical component in achieving broader sustainability objectives. Building on this foundation, Zhang et al. \cite{ZHANG2018251} conducted an in-depth analysis of energy consumption patterns in integrated steel plants, proposing a multi-objective optimization model that balances energy efficiency, production costs, and environmental impact. Their research highlights the intricate relationships between various production parameters and their collective influence on overall plant performance. In search of sustainability, Pardo and Moya \cite{PARDO2013113} explored the potential of breakthrough technologies in the steel industry to significantly reduce CO2 emissions. Their study evaluates emerging technologies such as hydrogen-based direct reduction and carbon capture and storage, providing insights into the future direction of sustainable steel production.

In addition to energy efficiency, improving operational quality and process control in steel production has garnered considerable attention. Kano and Yoshiaki explored methods to improve operational quality in steel manufacturing through advanced process control techniques \cite{kano2008data}. Their study underscores the relevance of optimizing production processes for quality improvement and resource efficiency, aligning with our objectives. In another effort, Zanoli, Cocchioni, and Pepe \cite{zanoli2018model} introduced a two-layer predictive control strategy based on linear models with adapted online horizons. This approach is applied within an advanced process control framework to optimize the reheating furnace of pusher-type billets in an Italian steel plant. The synthesized control system successfully replaced manual control actions, significantly improving process control, energy efficiency, and the fulfillment of product quality specifications. Further advancing process control, Branca et al. \cite{Branca2020} investigated the the current and future aspects of the digital transformation in the European steel industry. Their research demonstrates how digital transformation and smart manufacturing principles can revolutionize traditional steel production processes, leading to enhanced productivity and quality control.

DRL application in solving complex control problems has gained prominence in various domains. Mnih et al. \cite{mnih2015human} demonstrated the transformative potential of DRL by achieving human-level control in gaming environments. This landmark work underscores the adaptability and power of DRL in mastering intricate tasks, setting the stage for its exploration in industrial contexts. However, in industrial settings, the application of DRL is still evolving. Recent research has shown promising results. For example, Huang et al. \cite{Huang2019} applied DRL to optimize the scheduling of steel production processes, resulting in significant improvements in energy efficiency and production quality. Their work illustrates the growing interest in leveraging DRL for steel production optimization, mirroring our approach. Most recently, Ferreira Neto et al. \cite{FERREIRANETO2024110199} made significant achievements applying DRL to maintenance optimization in steel production. Their study proposes a DRL-based optimization approach for determining a scrap-based steel production line's optimal inspection and maintenance planning. The research considered practical aspects such as the uncertainty of maintenance duration and variable production rates, demonstrating the adaptability of DRL to complex, real-world industrial scenarios. Their findings indicate the potential for significant financial savings, highlighting DRL's power to enhance industrial competitiveness in the steel sector.

Although the incorporation of MLOps into industrial processes is an emerging and influential trend, its application remains relatively underexplored in heavy industries such as steel production. A growing body of technical reports highlights the importance and utility of MLOps in industrial contexts. However, comprehensive studies that showcase the results and lessons learned from implementing customized MLOps solutions are notably lacking. In particular, Zhao et al. \cite{zhao2022mlops} recently delved into the implementation of MLOps in the manufacturing sector, shedding light on its key role in ensuring the resilience and scalability of machine learning solutions within production environments. Their findings underscore the criticality of MLOps in effectively facilitating the operationalization of advanced ML techniques, aligning with the core objectives of our paper. However, despite these valuable insights, there remains a gap in the literature in presenting concrete results and experiences derived from applying customized MLOps solutions in heavy industries such as steel production. Building on this, Tao et al. \cite{Tao8477101} proposed a framework for integrating the concepts of MLOps with DT technology in smart manufacturing. Their research demonstrates how these concepts can enhance the accuracy and reliability of digital twin models, leading to more effective decision-making and process optimization in complex industrial settings like steel production.

Despite advances in applying ML and DRL to steel production optimization, significant challenges remain to implement and maintain these solutions in real-world industrial settings. Integrating MLOps and EDA in smart manufacturing presents both opportunities and hurdles, particularly in the steel industry. Faubel, Schmid, and Eichelberger \cite{Faubel2023} highlighted the critical role of MLOps in ensuring the resilience and scalability of ML solutions within production environments. However, applying MLOps in heavy industries such as steel production is still in its infancy. The challenge lies in adapting MLOps practices, typically used in software development, to the unique requirements of steel manufacturing processes. EDA offers the potential for real-time responsiveness in complex manufacturing environments. As noted by Theorin et al. \cite{Alfred2017}, EDAs can significantly enhance the flexibility and adaptability of control systems in industrial settings. However, integrating EDAs with existing steel production infrastructure, which often relies on legacy systems, presents considerable technical challenges. Combining MLOps and EDA in steel production control systems introduces new complexities in data management, model deployment, and system maintenance. As Tao et al. \cite{Tao8477101} observed in their work on digital twins in smart manufacturing, the sheer volume and variety of data generated in modern steel plants require sophisticated data processing and analysis pipelines. Ensuring the reliability and performance of these pipelines in a 24/7 production environment is a significant challenge.

\section{MLOps-Driven System Architecture for Digital Twin-based Control}\label{MLOPsDrivenSystemSEction3}

To automate the production process, reduce material waste, and improve production quality, we aim to develop AI-based software that can automatically control the power adjustments in the furnace induction coils based on data collected from the plant in real-time by the many different sensors. 

\textcolor{blue}{Our approach integrates three key components that will be discussed throughout this paper: (1) a Digital Twin that serves as a virtual model of the physical furnace, (2) an EDMA that enables real-time data processing and control, and (3) DRL algorithms that optimize decision making. In this section, we focus on the EDMA component, which forms the foundation for implementing and operating our DT. Later sections will demonstrate how this architecture supports the DRL algorithms that drive intelligent control decisions within the Digital Twin environment.}

To meet the stringent security requirements of the company, traditional off-the-shelf cloud solutions are not a feasible option. As temperature power adjustments need to be applied quickly due to the movement of the rod and the potential for overheating, an edge-computing approach is required. This requires a custom architecture framework to effectively manage and control production operations. To initiate this process, it is critical to establish a baseline production mode that can be automatically managed, potentially leveraging AI techniques, to ensure the system's correctness and efficiency. This operation state will be called the "proper normal production mode" throughout this paper.

For both normal production and warmholding modes, a robust operational infrastructure is required to facilitate AI-driven control algorithms in real-time production scenarios. While the AI algorithms that govern each production mode may differ according to specific conditions, the basic infrastructure remains constant. Consequently, our work focuses primarily on the development and implementation of the operational infrastructure, adopting an MLOps perspective. Our primary focus within this paper centers on automating the normal production mode. It is important to clarify that the emphasis here is not on the complexities of the algorithm, but rather on the architecture and infrastructure supporting it. Integrating a more complex algorithm into existing infrastructure follows a similar procedure, but such advanced algorithmic developments fall beyond the scope of this paper.

In our investigation of revolutionizing steel production by integrating DRL techniques, we have devised an MLOps architecture to guide our efforts. MLOps represents a structured framework necessary to manage the entire lifecycle of control software framework, including the ML solution. This framework is the backbone of our approach, which ensures seamless management of ML models throughout their lifecycle. The MLOps methodology comprises three key aspects: design, implementation, and deployment, each playing an important role in the success of our work.

In the design phase, we plan the entire transformation process. We assess the requirements for integrating DRL techniques and a DT into the steel production workflow, considering the unique challenges and complexities of this industry. We define the architecture, infrastructure, and workflow that will facilitate the seamless operation of our DRL-based system. During the implementation, the initial development of the ML model will take place, taking into account the data and the choice of the models. We also show the implementation of a robust ML pipeline that encompasses data collection, preprocessing, model training, validation, and evaluation. The heart of our implementation phase is the development of the DRL algorithm, which controls the electrical power consumption during the heating process. This phase involves the development and testing of the model to ensure the reliability and compatibility of the DRL model with the existing production infrastructure. We then show how to deploy the DRL model into the production environment and establish the infrastructure for real-time decision making. The following sections will go through these phases in detail and show the necessary parts of each phase.

The initial stage of the MLOps approach was the design phase. Here, we outline the overarching architecture and strategy to integrate DRL into steel production processes. To solve the problem described, we present the design of the system, which is composed of three parts:

\begin{itemize}

    \item \textbf{Furnace}, which can be considered just the production line, the sensors, and the actuators of interest;

    \item \textbf{Architecture for data input}, which collects data from plant sensors and makes them available for processing, together with receiving input to control actuators and request actions to be made;

    \item \textbf{Event-driven microservices architecture}, which performs a data processing.
    
\end{itemize}

\begin{figure}
    \centering
    \includegraphics[scale=0.99]{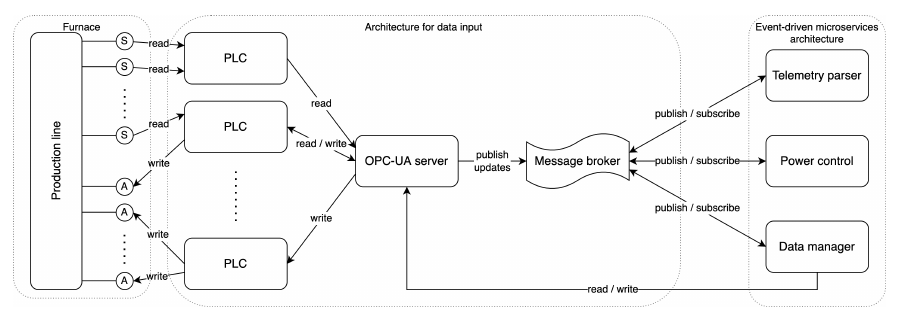}
    \caption{System architectural Overview}
    \label{fig:SystemArchitectureOverview}
\end{figure}

Figure \ref{fig:SystemArchitectureOverview} shows a high-level overview of these three parts of the system and how they are interconnected.

\subsection{Furnace}

Monitoring the forging process requires many sensors collecting data and a proper refresh interval to always have an up-to-date plant overview. Furthermore, it requires an adequate number of actuators for the operators to control the process's behavior. The furnace has sensors for different purposes placed in many parts to provide a complete overview of operations over time. The sensors measure the following:

\begin{itemize}
    \item The temperature, through pyrometers that are placed in every gap between induction coils and after shearing, to check if the billets have a sufficient quality to be used or need to be scrapped;

    \item The position of the rods inside the furnace;

    \item The velocity at which the roller conveyors move the rods;

    \item The power applied in the induction coils.

\end{itemize}

Operators can control, through the use of actuators, the following parameters in production:

\begin{itemize}
    \item the applied voltage in the induction coils, which defines the emitted power of the induction coils;
    \item the speed at which the roller conveyors turn, which defines the velocity at which the rods move;
    \item whether to let the furnace work in normal production or warmholding mode;
    \item whether to enable the automatic control or not.
\end{itemize}

\subsection{Architecture for data input}

All sensors and actuators are connected and controlled by Siemens 1500 PLCs (Programmable Logic Controllers). PLCs are ruggedized computer systems designed for industrial automation and control, capable of withstanding harsh industrial environments. The software of the PLCs is, in turn, connected to an instance of an OPC-UA\footnote{https://opcfoundation.org/about/opc-technologies/opc-ua/} server, which acts as a middleware layer between the data sources and the consumers. Within the OPC-UA server, the data are stored as strongly typed key-value pairs, called tags, which can be set to be read/written by OPC-UA clients. In addition, the PLCs receive data from the sensors and the OPC-UA server updates the values of the connected tags. Here, sanity checks can be performed by updating the values of the tags through code written for the PLCs. An OPC-UA server can work in two ways. In this paper, we use both methods in different situations:

\begin{itemize}
    \item \textit{client-server }or \textit{pull-based}, in which the tags’ values are requested from the server by the clients;
    
    \item \textit{publisher-subscriber} in which clients subscribe and receive updates every time a tag value is updated.
\end{itemize}

Instead of constantly accessing the OPC-UA server, tag updates are pushed on a topic/channel of a message broker (that is, Telemetry), which allows other services to subscribe to it. Figure \ref{fig:SystemArchitectureOverview} shows the connections between PLCs, OPC-UA, and the message broker.

\subsection{Event-driven microservices architecture}

The message broker topic contains the telemetry data coming from the OPC-UA server. This topic is the input for a software architecture that acts as a data processing pipeline, culminating in updates for the induction coils. These updates are voltages that, applied on the induction coils, would change the emitted powers. In this paper, we have selected the Event-Driven Microservices Architecture (EDMA) because the microservices have small tasks to fulfill and are designed to perform their jobs as fast as possible. Since there are many of them in the pipeline, low latency is a key requirement to avoid adding time overhead at every step. The \textit{event-driven} means that the data are processed mainly in a publisher / subscriber fashion (with a message broker acting as a middleware). The data that are pushed on the topics of the message broker is regarded as an event that must be processed. To this end, the data processing pipeline has the following steps:

\begin{itemize}
    \item The telemetry data gets parsed and reformatted to get stored and allow further analysis;
    
    \item The results get collected over time to form complex data structures, called state snapshots, which aim to represent the furnace's state at the state's creation time. These data structures also get stored for further analysis;
    
    \item from the reformatted telemetry, some pieces of data are either stored or cached in appropriate ways to be used within the processing of the pipeline;
    
    \item The state snapshots are given as input to some power control algorithms returning power updates (V), which are the voltages to set in the induction coils of every given zone;
    
    \item The power updates undergo sanity checks, and those voltages that pass them can be written to the corresponding tags in the OPC-UA server.
\end{itemize}

As is clear from Figure \ref{fig:SystemArchitectureOverview}, there are three main building blocks of the EDMA: 1) Telemetry Parser, 2) Power Control, and 3) Data Manager. The telemetry parser parses and reformats the telemetry data, creates state snapshots and tracks the material under production. The power control gets the state snapshots and processes them with power control algorithms that decide how to update the heating by the induction coils. The data manager gets instructions on how to update the heating of the induction coils, update the corresponding tags on the OPC-UA server, cache pieces of data from the OPC-UA server, and ensure that the connection with it is active. The following subsections illustrate the necessary details of these building blocks.

\subsubsection{Telemetry parser}

\begin{figure}
    \centering
    \includegraphics[scale=0.99]{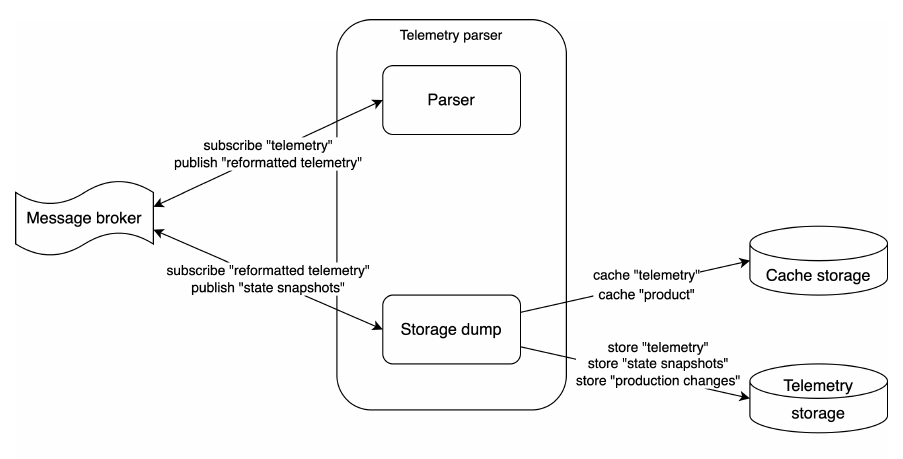}
    \caption{The Telemetry parser’s structure and its main interconnections}
    \label{fig:TelemetryParser}
\end{figure}

The telemetry parser is the macroservice in the data processing pipeline that processes the raw telemetry coming from the OPC-UA server. Figure \ref{fig:TelemetryParser} shows the microservices in place to perform the task of reformatting data and creating snapshots of the state.

As is clear from the structure, there are four active blocks in addition to the message broker: 

\begin{itemize}
    \item \textbf{Parser}, which gets the data from the message broker’s topic containing the telemetry, reformats it, and publishes it into another topic \textit{reformatted telemetry};

    \item \textbf{Storage dump}, which is subscribed to the \textit{reformatted telemetry} topic to store the incoming records in Telemetry storage, caches them together with data of the material under production in Cache storage, stores when the material under production changes, collects data to create state snapshots for later storing them in Telemetry storage and publishing them into a topic \textit{state snapshots};

    \item \textbf{Cache storage}, which is a storage service optimized to cache small amounts of data and has fast reading and writing times;
    
    \item \textbf{Telemetry storage}, is another storage service that has fast writing times and is optimized for memorizing structured and unstructured data.
    
\end{itemize}

\subsubsection{Power control}

\begin{figure}
    \centering
    \includegraphics[scale=0.99]{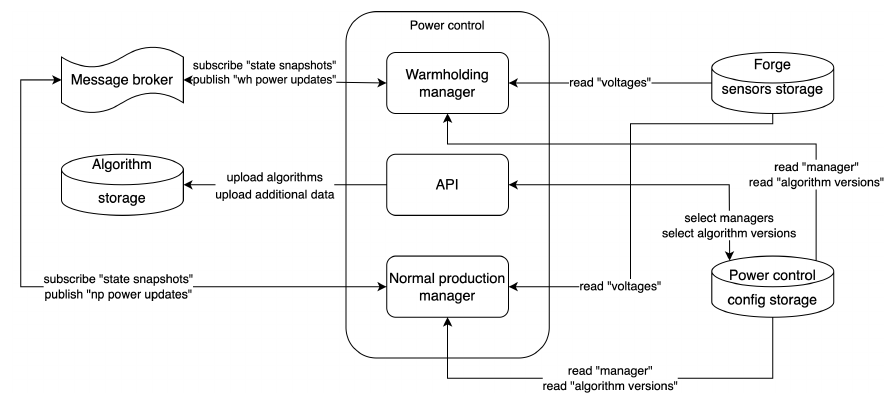}
    \caption{The Power control structure and its main interconnections}
    \label{fig:PowerControl}
\end{figure}

Power control is designed to consider the two modes that the furnace has to deal with: normal production and warmholding. For this reason, there are two microservices called \textbf{Normal production manager} and \textbf{Warmholding manager} that determine the power updates for the two different modes at any given time. Figure \ref{fig:PowerControl} shows the power control structure and the main interconnections. 

Both managers subscribe to the topic state snapshots and publish the voltages for the induction coils in the topic \textit{ np power updates} and \textit{wh power updates}. The former contains voltages applicable during normal production, the latter for warmholding. They are also updated with the current voltages applied to each zone by reading them from Forge sensors storage. Since the algorithms decide only what to do with the power (W) in the induction coils, there needs to be a way to convert from powers (W) to voltages (V). This conversion requires the current voltages for each zone, $V_{i,{ old }} \forall i \in 1, \ldots, 5$, read from the Forge sensors storage. The conversion also requires the power in each zone $P_{i,{ old }} \forall i \in 1, \ldots, 5$ before the algorithm makes its decision, based on the state snapshots. In addition, the new power $P_{i,{ new }}$ after the algorithm makes its decision is required for the conversion. Based on this, the new voltage for the $i^{th}$ zone is computed in Eq.\ref{Eq1}:

\begin{equation}\label{Eq1}
    V_{i, { new }}=\left\lceil V_{i, { old }} \cdot \sqrt{\frac{P_{i, { new }}}{P_{i, { old }}}}\right\rceil
\end{equation}

Therefore, a complete power update is computed in Eq.\ref{Eq2}:

\begin{equation}\label{Eq2}
     { PowerUpdate }=\left(V_{1, { new }} V_{2, { new }} V_{3, { new }} V_{4, { new }} V_{5, { new }}\right)
\end{equation}

More than one manager should ideally be available in the architecture since the microservice API allows for manager and algorithm selection, i.e., which manager to load for both the modes and which versions of the algorithms to use, as well as for uploading of new algorithm versions to Algorithm storage when they are ready. Rebooting is not necessary since the microservice is designed with a hot-swapping concept.

\subsubsection{Data manager}

\begin{figure}
    \centering
    \includegraphics[scale=0.99]{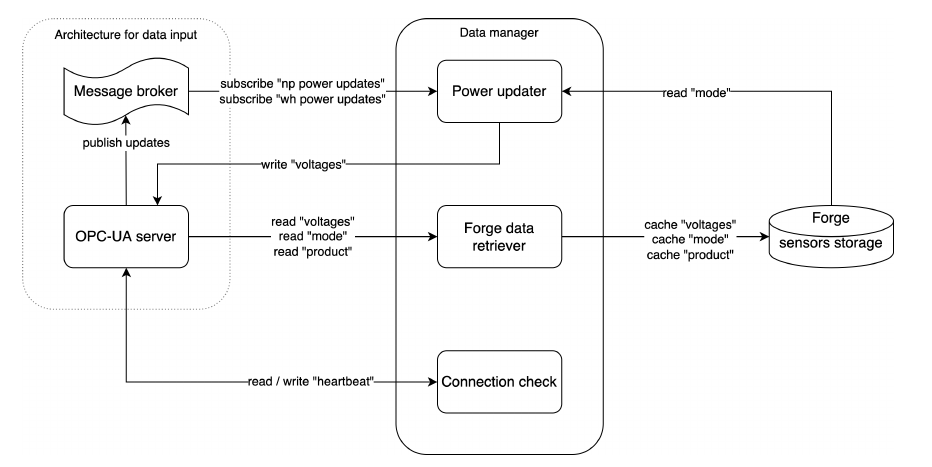}
    \caption{The Data manager’s structure and main interconnections}
    \label{fig:DataManagerStructure}
\end{figure}

The data manager is a collection of microservices that have only one of them connected to the message broker. Yet, all have communication links with the OPC-UA server. Figure \ref{fig:DataManagerStructure} shows this concept. The design idea is to concentrate on all microservices required to communicate with that server. Here, the power updater is subscribed to both the np power updates and wh power updates topics and updates the tags controlling the voltages of the induction coils on the OPC-UA server by selecting them from the topic corresponding to the currently active mode. The forge data retriever periodically retrieves the current voltages, the active mode, and material under production from the OPC-UA server and caches them into the forge sensors storage. The connection check constantly reads from and writes on a tag the OPC-UA server (as heartbeat) to check connection latency between the services and the server.

\section{DRL for Digital Twin (DT)-based Process Control}\label{DRLpowedControl}

\textcolor{blue}{Building upon the EDMA foundation described in the previous section, we now focus on how DRL is integrated with our DT to enable intelligent control of the steel production process. The DT environment, supported by our event-driven architecture, provides the ideal platform for implementing and training DRL algorithms that can learn optimal control strategies.}

\begin{figure}
    \centering
    \includegraphics[scale=0.9]{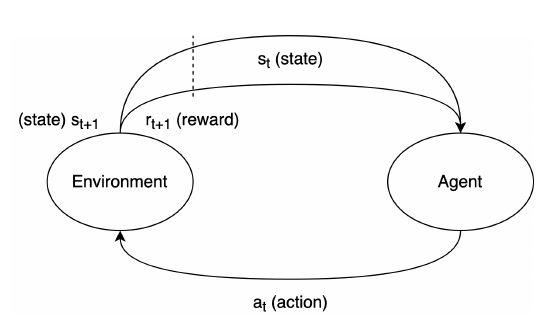}
    \caption{A high-level representation of the RL workflow}
    \label{fig:RLWorkFlow}
\end{figure}

In contrast to Supervised and Unsupervised ML, Reinforcement Learning (RL) is the approach to learning from interaction within an environment. The mathematical approach to RL is based on the theory of the Markov Decision Process (MDP), which allows describing problems of this type as tuples $\langle S, A, R, p, \gamma\rangle$ ($S$ is the set of possible states, $A$ is the set of performable actions, $R$ is the reward function, $p$ is the probability distribution that describes the next state. The reward given for the initial state and the action carried out in it $\gamma$ is the so-called discount factor) \cite{OGUNFOWORA2023244}. The learning entity in RL is called \textit{agent}, which interacts with the \textit{environment} to seek rewards to maximize them over time (the value to maximize is the expected cumulative reward, in which $R$ and $\gamma$ play an important role) \cite{sutton2018reinforcement}. Figure \ref{fig:RLWorkFlow} is a high-level representation of the RL workflow.

The agent can learn the best way to act through interaction with the environment. Taking into account a \textit{ model-free} RL problem, in which there is no access to the whole model, or it is impossible to have it, the approach to learning follows two paths:

\begin{itemize}
    \item \textit{policy-based} in which a policy $\pi$ is sought, which tells the agent what to do (i.e., what action to take) given a state $s \in S$ and an action $a \in A(\pi(a \mid s))$ is learned;

    \item \textit{value-based} in which a value function $Q_\pi$ or $V_\pi$ (depending on the input,\\ $\left.Q_\pi: S \times A \rightarrow \mathbb{R}, V_\pi: S \rightarrow \mathbb{R}\right)$ is learned, that implicitly use $\pi$, as remarked by the subscript in $Q$ and $V$.
\end{itemize}

The previous concepts of $\pi(a \mid s)$, $V(s)$ and $Q(s, a)$ are now represented by $\pi(a \mid s; \theta)$, $V(s; \theta)$ and $Q(s, a; \theta)$ in which $\theta$ are the weights of the DNN inside the agent \cite{Li17b}. The following sections explore one value-based method, namely Deep Q-Learning/Deep Q-Network (DQN) \cite{mnih2015human} and one policy-based method, Proximal Policy Optimization (PPO) \cite{schulman2017proximal}. DQN is chosen for its straightforward nature, popularity, and notable track record of delivering good results in the literature. In contrast, PPO is included to complement the evaluation, offering a complementary perspective to improve our understanding and provide a comprehensive assessment of the proposed approaches.

\subsection{Deep Q-Learning}

An important limitation of DQN is that actions can only be discrete. The goal is to approximate the optimal value function Q as shown in Eq.\ref{Eq.QLearning}:

\begin{equation}\label{Eq.QLearning}
    Q^*(s, a)=\max _\pi \mathbb{E}\left[\sum_{i=0}^{\infty} \gamma^i r_{t+1} \mid s_t=s, a_t=a, \pi\right]
\end{equation}

The experience replay is a memory from which the DRL agent can randomly sample the previous interactions with the environment and use them to update $\theta$ in the DNN representing the Q function. The target network is a DNN identical to the other one representing the Q function but is used to compute the expected cumulative reward. Its weights $\theta$ are not updated together with $\theta$; instead, only every $C$ step ($C$ is a hyperparameter of the training process). The actions are usually chosen with a $\epsilon$-greedy strategy to balance exploration and exploitation.

\subsection{Proximal Policy Optimization (PPO)}

The idea behind PPO is to try to optimize a similar algorithm, namely Trust Region Policy Optimization (TRPO), and combine it with an Actor-Critic with Experience
Replay (ACER) approach. The actor network is trained to learn the policy $\pi$, while the critic network tries to learn the value function $V$. To learn the optimal $\pi$,
the agent tries to minimize the loss:

\begin{equation}
    L_t^{C L I P+V F+S}(\theta)=\hat{\mathbb{E}}\left[L_t^{C L I P}(\theta)-c_1 L_t^{V F}(\theta)+c_2 S\left[\pi_\theta\right]\left(s_t\right)\right]
\end{equation}

in which:

\begin{equation}
    L_t^{C L I P}(\theta)=\hat{\mathbb{E}}\left[\min \left(r_t(\theta) \hat{A}_t, \operatorname{clip}\left(r_t(\theta), 1-\varepsilon, 1+\varepsilon\right) \hat{A}_t\right)\right]
\end{equation}

where $r_t(\theta)=\frac{\pi_\theta\left(a_t \mid s_t\right)}{\pi_{\theta_{\text {old }}}\left(a_t \mid s_t\right)}$ represents the ratio between the new and old policies at each step. The loss $L_t^{C L I P}(\theta)$, as shown, uses a combination of min and a clipping function to limit this ratio: the idea of doing so is to limit aggressive positive updates to $\pi$ (to converge slower but more stable) and instead allow negative updates, when necessary. The $\hat{A}_t$ is the advantage at time t, calculated as $\hat{A}_t=\delta+\gamma \cdot \lambda \cdot m_t \cdot \hat{A}_{t+1}$, with $\delta=r_t+\gamma \cdot V\left(s_{t+1}\right) \cdot m_t-V\left(s_t\right)$ and $m_t$ a mask value of 1 when a transition completed an episode. Finally, $L_t^{V F}(\theta)=M S E\left(V_\theta\left(s_t\right), V_t^{\text {targ }}\right)$. To balance exploration and exploitation, the output of the policy in a given state is used as a multinomial probability distribution and the action is sampled with it.

The system is separated into two parts:

\begin{itemize}
    \item \textbf{Infrastructure Part and Data source: }it comprises the industrial plant, the sensors, the PLCs and the OPC-UA server, located at Bharat Forge Kilsta;
    \item  \textbf{Data streaming and processing:} includes a message broker that acts as a source for the EDMA and the EDMA itself, which can be placed on an edge compute platform. 
\end{itemize}

For the Infrastructure and Data source part, the PLCs that control the furnace export sensor telemetry data to the local OPC-UA server, which can also manipulate the PLC by writing commands into it. By connecting our data streaming and processing functionality to the OPC-UA server, we thus implicitely can control the PLC through writing to specific tags in the OPC-UA server. In the following subsection, we summarize how the detailed components of the system were implemented, focussing on the Data streaming and processing parts. 

\subsection{Storage services}

Considering the choice of storage service, it is crucial to carefully assess the specific requirements and determine the most appropriate technology to meet those needs. As mentioned previously, the following storages are demanded by the microservices:

\begin{itemize}
    \item \textbf{Cache storage} to store small pieces of data from raw telemetry and the current material in production;

    \item \textbf{Telemetry storage} to store all telemetry records, the snapshots created of the state, and the production changes that occurred;

    \item \textbf{Algorithm storage} to store algorithms (binaries, scripts, or plain text files to be processed) and additional data that could be helpful for later use;

    \item \textbf{Power control config storage} to store which manager is active for each of the two production modes and which versions of the algorithms are loaded;

    \item \textbf{Forge sensors storage} to store the current production mode, the current demanded voltages in each furnace zone, and the current material in production.

\end{itemize}

To represent those storages in a systematic way, three techniques were used for storage: 

\begin{itemize}
    \item \textbf{key-value} for an easy setup, fast reading, and writing of small pieces of data. For this purpose, Redis [31] (version 6.2.6) - which works in memory and therefore is extremely fast - is used;

    \item \textbf{NoSQL for structured/semi-structured data} to support storing both structured and unstructured data, focusing on writing speed and the possibility of horizontal scaling in case of need. Apache Cassandra [1] (version 4.0.3) was chosen because of its robustness and ability to scale without trouble;

    \item \textbf{file system} for storing files, binaries, and scripts. Nothing special was used in this case, just the plain file system where the software runs.

\end{itemize}

An instance of Redis named forge\_sensors\_storage is therefore reachable by all the microservices in the architecture and comprises the cache storage, power control configuration storage, and Forge sensors storage that have just been presented. Instead, the Cassandra instance maps only to Telemetry storage and is named after it: telemetry\_storage. Finally, the Algorithm storage uses the file system.

\subsection{Message broker}

The characteristics of a message broker to allow communication among microservices to allow EDA to function well are high throughput, low latency, and the possibility of having multiple topics. For the architecture presented, it is necessary that this broker temporarily stores the messages that are sent so that if a service crashes or reboots, the data are not lost. The broker can scale horizontally when needed and the throughput is high enough to support all telemetry data to be sent almost at the same time on the same topic (about 200 records per second). Here, the writing and reading time is low to allow almost real-time data processing (in particular, it has to be possible to process all telemetry sent during the same second to try creating a state snapshot). The broker also maintains the order of the messages sent, at least under certain conditions. 

All these characteristics are respected by the chosen message broker, Apache Kafka\footnote{Apache Kafka. URL: https://kafka.apache.org/.} (version 5.5.3 released and maintained by Confluent Inc.), which also has the good aspect of being very efficient and reliable. Since all microservices are written in Python to have a coherent technology stack, the library used to connect to Kafka is called Faust\footnote{Faust. URL: https://faust.readthedocs.io/en/latest/.}, which is also one of the few libraries that supports the Kafka Streams API in Python.

\subsection{Integration with the plant (OPC-UA)}

Updates to OPC-UA tags will be pushed into the message broker's (i.e., Kafka) topic telemetry. The creation of this stream of data is allowed by combining different tools:

\begin{itemize}
    \item the connection to OPC-UA is performed through a ThingsBoard IoT Gateway\footnote{ThingsBoard IoT Gateway. URL: https://thingsboard.io/docs/iotgateway/what-is-iot-gateway/.} that offers a straightforward setup and an OPC-UA connector and data visualization out-of-the-box;

    \item the data collected through the OPC-UA connector is converted to a preliminary JSON object and pushed to an MQTT broker (Eclipse Mosquitto\footnote{Eclipse Mosquitto. URL: https://mosquitto.org/.}) to transform it into a stream of data in the publisher / subscriber fashion;

    \item a Kafka connector for MQTT is subscribed to the MQTT topic and eventually pushes the JSON objects as messages in a Kafka topic named v1\_gateway\_telemetry.
\end{itemize}

\begin{figure}
    \centering
    \includegraphics[scale=0.99]{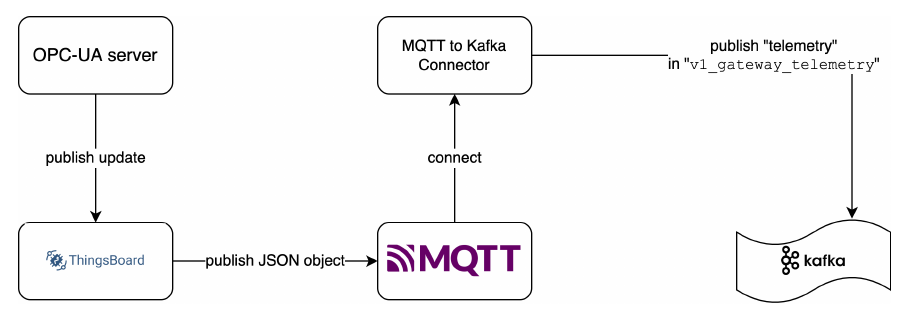}
    \caption{How tools communicate for integrating OPC-UA with the Kafka instance}
    \label{fig:OPCUAcommunicate}
\end{figure}

Figure \ref{fig:OPCUAcommunicate} shows how these components and tools communicate to integrate OPC-UA with the Kafka instance.

\subsection{DRL Agent and algorithm structure}\label{section4.6}

The algorithm for training follows the structure of a classic RL/DRL problem, with differences in how the inner logic of the DRL agent is designed, which clearly depends on whether it is using PPO or DQN. The essential parts of it are presented through pseudocode in Algorithm \ref{alg:DRL-Train}. The syntax of the environment usage resembles that of the Python library OpenAI Gym\footnote{OpenAI Gym. url: https://www.gymlibrary.dev/.} requires.

\begin{algorithm}
    \centering
    \caption{DRL-Train($env$, $n\_actions$, $state\_size$, $episodes$, $learning\_rate$, $batch\_size$, $config$)}

    \includegraphics[scale=0.9]{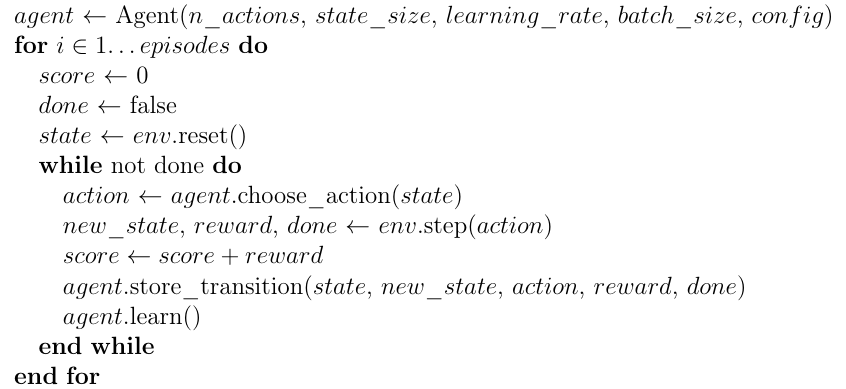}
    \label{alg:DRL-Train}
\end{algorithm}

\begin{figure}
    \centering
    \includegraphics[scale=0.9]{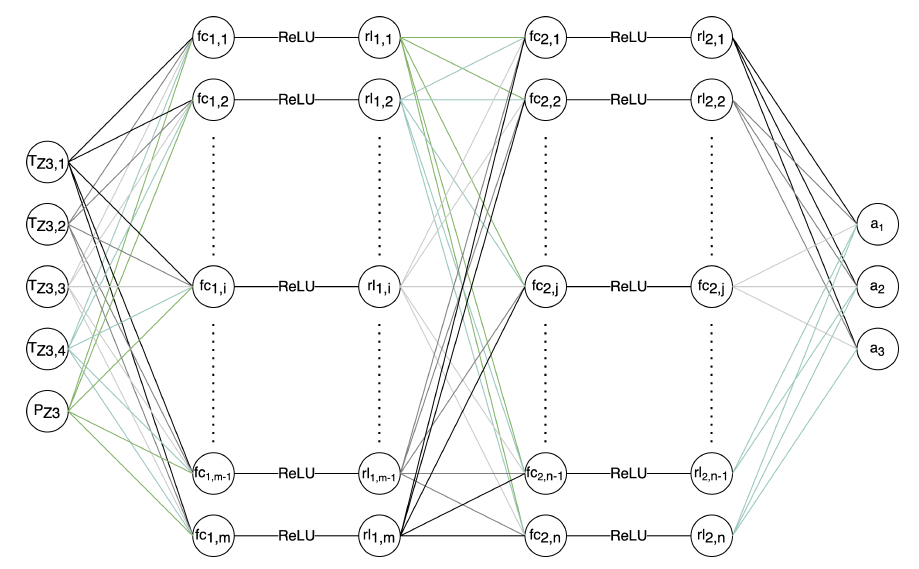}
    \caption{The DQN graphically represented (assuming the forge sensors were used).}
    \label{fig:DQNStructure}
\end{figure}

Figure \ref{fig:DQNStructure} shows a graphical representation of the Deep Q-network and Target network. \textcolor{blue}{The input layer has} four $T_{(Z_{(3,1)}}, ...., T_{(Z_{(3,4)})}$ or 15 neurons $T_{(Z_{(3,1)}}, ...., T_{(Z_{(3,15)})}$, depending on whether the simulator is configured to use the forge or virtual sensors, for the temperatures registered in the third zone, and 1 for its power ($P_{Z3}$). Two fully connected, hidden layers have $m$ and $n$ neurons, followed by a ReLU activation function. The output layer is another fully connected layer and has three output neurons ($a_1, a_2, a_3$), each of them representing one action the agent can take (the output of the model is the raw value of the neuron without activation functions such as softmax).

Given A = ($a_1, a_2, a_3$) the output of the neural network, the action $a^*$ the agent
would take is:

\begin{equation}
    \binom{a^{*}=argmax (a_{i})-1}{i\in 1,2,3}
\end{equation}

When training, $a^*$ is chosen only with probability $1-\varepsilon$, as defined by the $\varepsilon-greedy$ strategy.

\begin{figure}
    \centering
    \includegraphics[width=0.9\linewidth]{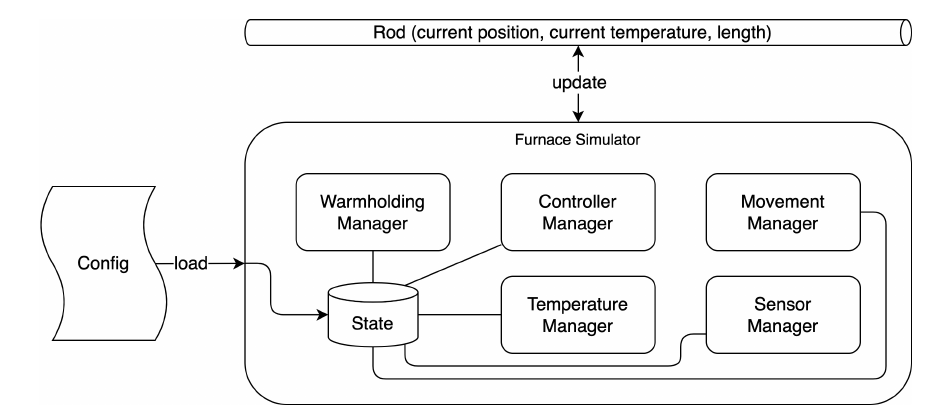}
    \caption{A graphical representation of the furnace DT}
    \label{fig:DTGraph}
\end{figure}

The DT is designed to replicate the functionality of the actual furnace while simultaneously offering insights beyond those available from sensor data in the physical environment. Figure \ref{fig:DTGraph} graphically represents the furnace DR.

From Figure \ref{fig:DTGraph}, several aspects can be discerned. The DT is initialized by loading a configuration. This includes setting parameters such as the duration of a simulation step (in seconds), the total number of simulation steps, the initial velocity of the rod (in meters per second), the starting power of the induction coils (in kilowatts), the placement of sensors and the specific start and end points of each coil (in meters, relative to the warehouse's starting point), and the ambient room temperature (in degrees Celsius). Once configured, these settings are integrated into the DT's state. The Managers are linked to the DT state, which is then updated throughout the simulation. The rod, along with the configuration, serves as a real-time input to the DT. Its position and temperature are updated at each step in real-time. Here, the Warmholding Manager monitors the status of warmholding and acts accordingly. The Controller Manager adjusts the power of the induction coils. The Movement Manager updates the position of the rod, and the Temperature Manager modifies the rod's temperature. The Sensor Manager records the temperature at each sensor location.

\begin{figure}
    \centering
    \includegraphics[width=0.75\linewidth]{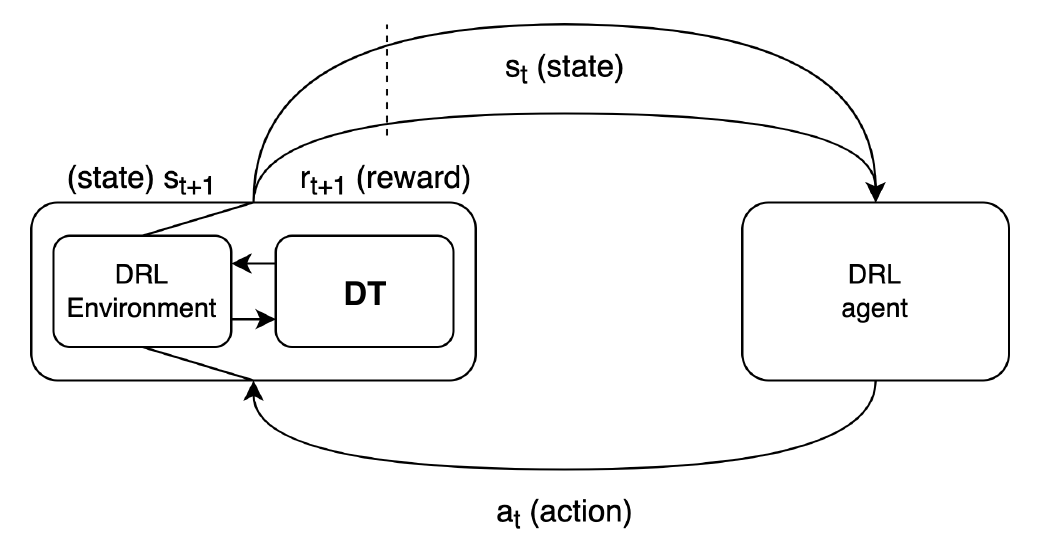}
    \caption{The DRL cycle for the problem and its integration with the DT}
    \label{fig:DRLwithDT}
\end{figure}

The DT is capable of operating over single or multiple steps. The distinction between these two modes is significant: operating step-by-step allows for control over the process at each individual time step, whereas running it over multiple steps provides control only at the conclusion of the entire process. This distinction is particularly relevant for the DRL agent, which requires the ability to manage events within the DRL environment at its own pace. Consequently, the DT is designed to accommodate various implementations of managers. In alignment with this requirement, the DRL agent typically performs the simulation for a single step, during which it evaluates the rod's position and temperature profile. Once this analysis is completed, the agent starts the next step of the simulation, using the output of the rod from the previous step as its starting point. As mentioned previously, the DRL is built on top of the DT. The DT is fed with real-time production data. Figure \ref{fig:DRLwithDT} shows the DRL cycle for the problem and its integration with the DT.

The detailed implementation of the DT, its internal structure, and its evaluation are beyond the scope of this paper. More details about the design and implementation of the DT can be found in our earlier publication \cite{Ma2022UsingDR}. DR was extensively evaluated and tested in our earlier study, which can be found in \cite{YunpengTesting2023}.

\section{Deployment}\label{Deployment}

\textcolor{blue}{The deployment of our DT solution, with its DRL algorithms running within the EDMA framework, presents unique challenges and requirements. In this section, we focus on the MLOps aspects of deploying, maintaining, and updating DRL models within our DT environment, ensuring that the virtual representation remains synchronized with the physical furnace.}

Model maintenance and model management updates are crucial in MLOps. With the availability of many models, only one Power Control Manager loads them. To this end, the exported models must have a shared interface to keep a level of model abstraction and allow the manager to interface with so many models' inner details. When one model is ready to be deployed, it must be wrapped in advance with layers that translate the common input to the input of the model and the output to the common output. The interface is unified between the DRL models. Its design tries to take into account all possible training needs. Since the plant has 18 temperature sensors and five power values for the five zones, the model has 23 input features. These 23 features include 18 temperature values (2 for zone 1 and 4 for zones 2 to 5) in °C, followed by five power values in kW. The interface will also take the recorded temperatures and expected powers from the state snapshots. It also takes 20 output features that are five (logistic) probability distributions representing, for each zone in ascending order, the probability of increasing, decreasing, not changing, or dropping to a very low value the power applied to each zone.

The inner logic of the exported model will also follow a unified mechanism during deployment. It removes any input feature that is not necessary and normalizes the input features in [0, 1] when needed. It also finds a function $f$ that interpolates the temperatures of the forge sensors at the positions of the forge sensors to compute the temperatures the virtual sensors would register. Then, it builds the output for the required zones and sets to zero the remaining unused values.

\begin{figure}
    \centering
    \includegraphics[scale=0.7]{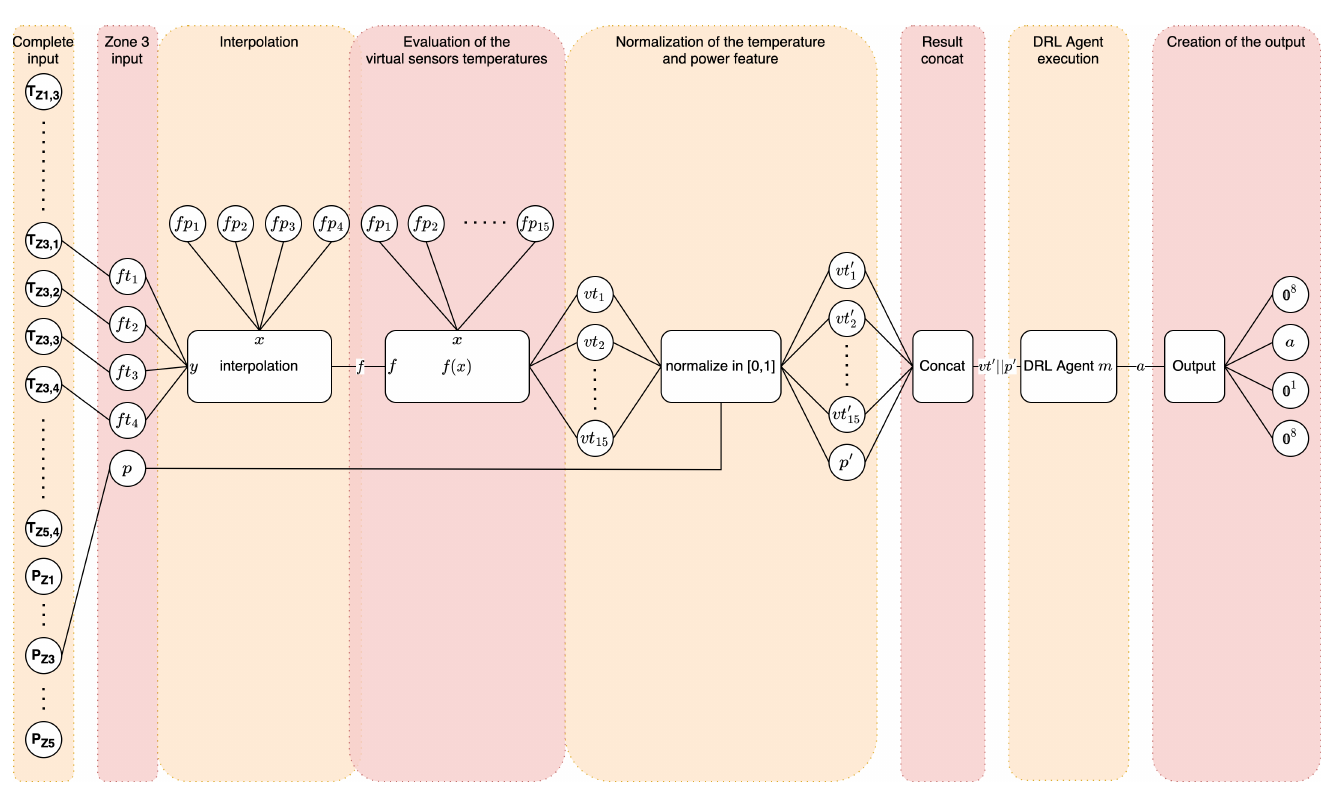}
    \caption{Graphical representation of an example of a model wrapper and deployment.}
    \label{fig:ModelWrapper}
\end{figure}

For any input $x$, the original model $m$ and the exported model $\acute{m}$ should always return the same value, that is, $m(x) =\acute{m}(x))$. In the case of a DRL agent $m$ controlling zone 3 that normalizes the input and uses virtual sensors, exporting it means creating a wrapper $\acute{m}$. The wrapper $\acute{m}$ keeps the input features from the 6th to the 10th and the 21st (temperatures and power in zone 3) and finds an interpolation function $f$ $s.t.$ $\forall i \in 1,...,4 f(fp_i)=ft_i $, considering $ft_i \forall i \in 1,...,4 $ as the four temperature features and $fp$ the positions of the corresponding forge sensors. The wrapper also computes the temperatures $vt_i=f(vo_i) \forall i\in 1,...,15$ that would be recorded by the virtual sensors at each of their positions $vp_i$. It is also necessary for the model wrapper to normalize $vt$ and the value of the power feature $p$ in the range [0, 1], resulting in $v\acute{t}$ and $\acute{p}$, and also calculate $a=m(v\acute{t}||\acute{p})$ with $a\in {\mathbb R}^3$ corresponding to three possible actions. Figure \ref{fig:ModelWrapper} shows a graphical representation of the wrapper behavior.

\section{Experimental Evaluation}\label{Evaluation}

\textcolor{blue}{In this section, we evaluate our integrated system that combines the DT, Event-Driven Microservices Architecture, and DRL algorithms. While our evaluation focuses primarily on the performance of the DRL algorithms, it is important to note that these results are achieved within the context of our DT environment, which is enabled by the EDMA infrastructure described earlier.}

Here, we aim to conduct an empirical assessment to evaluate the operational performance and compliance of the Event-Driven Microservices Architecture (EDMA). The primary objectives of this assessment are to determine the extent to which the EDMA aligns with its intended functionality, ensure that the microservices implementations adhere to the specified design requirements, and verify that they operate within expected timeframes. 

The experimental data presented in this section were gathered using a computing environment with the following hardware and software specifications:

\begin{itemize}
    \item Operating system: Ubuntu 20.04.1 LTS (GNU/Linux 5.4.0-124-generic x86\_64)
    \item CPU: Intel Xeon Silver 4110 CPU @ 2.10GHz (8 cores)
    \item RAM: 46 GB
    \item Docker: 20.10.12, build e91ed57
\end{itemize}

Our investigation aims to assess whether the EDMA functions as intended by examining the microservices' correct behavior and performance within this specific computational environment.

\subsection{Processing Time: Maximum Expected vs. Actual Average}

To ensure the correctness and effectiveness of microservices, it is imperative to assess their actual performance against their expected processing times. This evaluation focuses on explaining the specified time windows within which each microservice is intended to run. For the three main microservices, it is important that the microservices comprising the Telemetry Parser effectively process incoming sensor data, which may reach up to 200 records per second. Consequently, it is crucial that each sensor's data, including the processing required for contributing to state snapshots, be handled within a maximum time frame of 5 milliseconds. For the Power Control module, particularly the DRL Normal Production Manager, it is necessary to process approximately one snapshot per second. Within the Data Manager component, the Power Updater must process approximately one power update per second. This ensures that the plant's operational parameters are promptly updated on the basis of recent snapshots. In addition, it is imperative to maintain up-to-date information on the current voltages applied to the induction coils and to update the active production mode in response to any changes.

The processing time thresholds for each microservice were determined based on the operational requirements of the steel production process and the technical constraints of our system:

\begin{enumerate}

    \item \textcolor{blue}{ \textbf{Telemetry Parser (5 ms threshold):} This component must rapidly parse incoming sensor telemetry and organize it for further processing. The threshold ensures smooth data flow without backlog buildup.}
    
    \item \textbf{Power Control (1000 ms threshold):} The induction furnace's thermal dynamics dictate this threshold. Our process engineers at Bharat Forge Kilsta determined that updating power settings more frequently than once per second would not significantly improve control but could potentially cause system instability.
    
    \item \textbf{Data Manager (1000 ms threshold):} This aligns with the Power Control threshold to ensure that each power update is promptly communicated to the OPC-UA server, maintaining synchronization between our control system and the physical furnace.
\end{enumerate}

These thresholds represent the maximum allowable processing times to maintain real-time control and system stability in the steel production environment.

The actual processing times have been derived from an extensive collection of data spanning two to three months, obtained through comprehensive log analysis, ensuring the reliability and accuracy of the presented results. These empirical findings are detailed in Table \ref{tab:processing-times}.

\begin{table}[ht]
\centering
\caption{Comparison of Maximum Expected and Actual Processing Times}

\begin{tabular}{|c|c|c|}
\hline
\textbf{Microservice} & \textbf{Max proc. time (ms)} & \textbf{Actual proc. time (ms)} \\
\hline
Telemetry parser & 5 & 4.622 \\
\hline
Power control & 1000 & 4.055 \\
\hline
Data manager & 1000 & 9.922 \\
\hline
\end{tabular}
\label{tab:processing-times}
\end{table}

As can be seen in the table, the processing time for the Telemetry Parser begins when events are published on Kafka. Since this microservice does not interact directly with external systems like OPC-UA, its processing time depends mainly on internal operations and Kafka message handling. It may not fully capture the network-related latency between the OPC-UA server and the EDMA. The Power Control microservice performs remarkably efficiently, even considering the network-related latency. Despite potential delays in data retrieval or communication with external systems, it still manages to process snapshots well within the expected time frame. This suggests that it effectively handles any inherent latency challenges. The higher average processing time for the Data Manager, especially the Power Updater microservice, can be attributed to its need to connect to OPC-UA to send power updates. The network latency introduced by the VPN connection likely contributes significantly to this increased processing time. Despite having less computational work to perform, the impact of network delays becomes more pronounced in this microservice.

\subsection{DRL power control algorithm}\label{section9.2}

In this section, we present the results obtained from a series of training jobs conducted for the power control algorithm within Zone 3 of the furnace. We chose to focus on Zone 3 for several key reasons. Firstly, Zone 3 operates in a critical temperature range (1140°C to 1275°C, as shown in Figure \ref{fig:TempratureRange}) where precise control is crucial for product quality. Secondly, this zone represents a transition point between the intensive heating of the first two zones and the final temperature stabilization in the last two zones, making it particularly challenging to control. Moreover, the control challenges in Zone 3 are representative of the overall furnace control problem, providing insights that can be generalized to other zones. Additionally, given the computational intensity of DRL training, focusing on one zone allowed for a more thorough exploration of hyperparameters and algorithms within our available resources. Although we conducted experiments for all zones, the results from Zone 3 provide a comprehensive view of our DRL approach's effectiveness. The insights gained from this zone can be applied to optimize control strategies across the entire furnace.

These experiments included a wide range of hyperparameter configurations, variations in the loading of the DRL environment, and the reward function used. The primary objectives of this section are to determine the optimal algorithm for various scenarios, quantify the reduction in execution time achieved by using the simulator environment, and identify reward functions that produce improved temperature profiles. The best model will then be deployed for inference.

The experiments were conducted on the "Berzelius" compute cluster, administered by the National Supercomputer Center (NSC) at Linköping University, Sweden. The node specifications for this cluster are as follows:

\begin{itemize}
    \item Operating system: RHEL 8 (4.18.0-348.20.1.el8\_5.x86\_64)
    \item CPU: AMD EPYC 7742 64-Core Processor (16/64 cores assigned)
    \item RAM: 125/1000 GB assigned
    \item GPU: NVIDIA A100-SXM4-40GB
    \item CUDA drivers: 11.2
    \item Python: 3.10.4
    \item PyTorch: 1.12.0+cu113
\end{itemize}

Subsequent analysis in the following subsections will dive into the findings, highlighting the performance differences between algorithms, the time reductions afforded by the DT, and the impact of reward functions on achieving optimal temperature profiles in Zone 3.

\subsubsection{Collected metrics}

In each training job, we systematically collect various metrics about both the performance of the algorithm, including the "status" of the DNN, and the outcomes generated within the DRL environment. These metrics encompass the following data points for each experiment:

\begin{itemize}
    \item \textbf{Selected Hyperparameters:} The configuration parameters chosen for the experiment are fundamental to influence the behavior and learning of the algorithm.

    \item \textbf{Episode Score:} This metric represents the cumulative rewards acquired during each episode. It provides information on how well the agent achieves its objectives.

    \item \textbf{Mean Loss per Episode:} The mean loss value computed for each episode. This calculation aggregates losses incurred during the episode, reflecting the algorithm's convergence or divergence behavior.

    \item \textbf{Mean Step Time per Episode:} The average time taken per step within an episode. It helps gauge the temporal efficiency of the algorithm's decision-making process.

    \item \textbf{Total Episode Time:} The cumulative duration of each episode indicates how long the agent took to complete its decision-making process within the environment.

    \item \textbf{Number of Steps per Episode: }The count of discrete steps taken within each episode. This number is typically fixed according to the current rod configurations and offers insights into the depth of exploration the agent engages in.

    \item \textbf{Temperature at the Last Sensor in Zone 3 (with every 500 steps): } The temperature recorded by the final virtual sensor in Zone 3, sampled at regular intervals of 500 steps. These data provide an ongoing snapshot of the temperature dynamics within the controlled zone.

    \item \textbf{Power Applied to Induction Coils in Zone 3 (Every 500 steps):} The magnitude of the power applied to the induction coils in Zone 3, also measured at regular intervals of 500 steps. This information is crucial to understanding the control actions executed by the agent.

    \item \textbf{Epsilon Value (DQN only):} For the $\epsilon{-greedy}$ strategy used by the DQN algorithm, we monitor the current value of $\epsilon$, which influences the exploration-exploitation trade-off. This value guides the agent's propensity to take random actions versus exploiting known policies.
\end{itemize}

Collecting and analyzing these metrics across various experiments allows a comprehensive assessment of algorithmic performance, hyperparameter sensitivity, convergence behavior, and the impact of chosen configurations on the DNN and DRL environment.

\subsubsection{Hyperparameters and Episodic Score}

Each training job is initiated with a unique combination of hyperparameters primarily dependent on the choice of algorithm (DQN or PPO). However, certain hyperparameters are shared across all jobs to ensure consistency in the evaluation process. During the experiments, we cover the "Number of Episodes" to determine the number of training episodes for each job. To specify the learning rate of the optimizer used during training, we use the "Learning Rate" parameter. We also use the "Random Seed" parameter for libraries that generate pseudorandom numbers to ensure reproducibility. To dictate the number of transitions selected from memory at each training step, we use the "Batch Size" parameter. We use "Normalize Observations", which is a Boolean flag that indicates whether observations should be normalized during training. We also used a Boolean flag that determines whether noise should be applied to the power levels of the first two zones and a Boolean flag that allows for the choice between virtual or forge sensors for data acquisition.

Table \ref{tab:common-hyperparameters} presents the permissible values that each of these common hyperparameters can assume during a training job, providing information on the flexibility and configurability of the training process.

\begin{table}
\caption{Common Hyperparameter Values}
\centering

\begin{tabular}{|c|c|}
\hline
\textbf{Hyperparameter} & \textbf{Values} \\
\hline
Episodes & 50, 100, 200, 300 \\
\hline
Learning rate & 0.001, 0.0001, 0.00001 \\
\hline
Seed & 19, 39 \\
\hline
Batch size & 64, 128, 256, 512, 1024, 2048, 4096 \\
\hline
Normalization & Enabled (True), Disabled (False) \\
\hline
Z1 and Z2 powers & With noise (False), Without noise (True) \\
\hline
Sensors & Virtual (False), Forge (True) \\
\hline
\end{tabular}
\label{tab:common-hyperparameters}
\end{table}

For the DQN algorithm, a distinct set of hyperparameters is customized to optimize its performance. These hyperparameters are essential for shaping the behavior and learning process of the DQN agent. We used \textbf{$\gamma$ (Gamma)} that determines the discount factor for the Bellman equation, which influences the agent's preference for immediate rewards versus long-term rewards. \textbf{Starting $\epsilon$ (Epsilon)} is used to signify the initial value of the $\epsilon$-greedy exploration strategy. It regulates the agent's propensity to explore the environment versus exploiting its current knowledge. \textbf{Lower Bound for $\epsilon$} is also used to set the minimum value that $\epsilon$ can reach during the training process. The \textbf{Decrease Step for $\epsilon$} is used to specify the rate at which $\epsilon$ is reduced over time. \textbf{DNN's First Fully Connected Layer Neurons} is used to determine the number of neurons in the first fully connected layer of the Deep Neural Network (DNN) architecture used to estimate Q values. Similarly, \textbf{DNN's Second Fully Connected Layer Neurons} is used to determine the number of neurons in the second fully connected layer of the DNN. The DQN algorithm employs a target network to stabilize the training. The \textbf{Number of Steps Before Updating Target Network Weights $\theta'$} defines the frequency at which the weights of the target network $\theta'$ are updated. Finally, the \textbf{Transitions Memory Size (Replay Memory Size)} parameter is used to define the size of transition memory (often referred to as replay memory), which stores experiences for replay during training.

These DQN-specific hyperparameters are critical in configuring the DQN agent's learning dynamics and overall training performance, allowing for fine-tuned adjustments to address specific task requirements and challenges. Table \ref{tab:dqn-hyperparameters} shows the values each DQN hyperparameter can assume during a job.

\begin{table}
\centering
\caption{DQN Hyperparameter Values}

\begin{tabular}{|c|c|}
\hline
\textbf{Hyperparameter} & \textbf{Values} \\
\hline
$\gamma$ & $0.9, 0.95, 0.98, 0.99$ \\
\hline
$\epsilon$ in $\epsilon$-greedy & $0.7, 0.8, 0.9, 1.0$ \\
\hline
$\epsilon$'s lower bound & $0.01, 0.001, 0.0001, 0.00001$ \\
\hline
$\epsilon$'s decrease step & $0.05, 0.005, 0.0005, 0.00005$ \\
\hline
Fully connected 1 neurons & $128, 256, 512$ \\
\hline
Fully connected 2 neurons & $128, 256, 512$ \\
\hline
Target network update interval & $1000, 10000, 100000$ \\
\hline
Transitions memory & $100000, 200000, 500000$ \\
\hline
\end{tabular}
\label{tab:dqn-hyperparameters}
\end{table}

In the context of the PPO algorithm, several hyperparameters come into play to fine-tune and customize the training process. The $\lambda$ (Lambda) Value represents the weight applied to the Generalized Advantage Estimate (GAE). The coefficient $c_1$ plays an important role in shaping the PPO loss function. The $\epsilon$ (epsilon) determines the extent to which policy updates are constrained, as detailed in the aforementioned section. The number of training epochs executed during each training round that influences the convergence and stability of the learning process. The Number of Steps Before Training defines the number of steps taken before retraining the Actor and Critic networks and the size of the memory of the transition. The Actor's First Fully Connected Layer Neurons parameter is used to specify the number of neurons in the first fully connected layer of the Actor-network affects the network's capacity to represent policy functions. Similarly to the first layer, the number of neurons in the second fully connected layer of the actor network influences the policy function's expressive power. Table \ref{tab:PPO_hyperparameters} shows the values that each PPO hyperparameter can assume during a job.

\begin{table}
\centering
\caption{Values PPO hyperparameters }
\begin{tabular}{|c|c|}
\hline
\textbf{Hyperparameter} & \textbf{Values} \\
\hline
$\lambda$ for GAE & $0.9, 0.95, 0.98, 0.99, 1.00$ \\
\hline
$c_1$ in loss & $0.5, 1.0$ \\
\hline
$\epsilon$ in the policy clip & $0.1, 0.2, 0.3$ \\
\hline
Epochs & $5, 10, 15, 20, 25, 30$ \\
\hline
Training interval & $10, 25, 50, 100, 200, 500, 1000, 2000, 5000$ \\
\hline
Actor fully connected 1 neurons & $128, 256, 512$ \\
\hline
Actor fully connected to 2 neurons & $128, 256, 512$ \\
\hline
Critic fully connected 1 neurons & $128, 256, 512$ \\
\hline
Critic fully connected 2 neurons & $128, 256, 512$ \\
\hline
\end{tabular}
\label{tab:PPO_hyperparameters}
\end{table}

The results presented have been obtained through the implementation of DRL agents for the two aforementioned scenarios of normal production. We conducted a comprehensive hyperparameter search, systematically exploring the combinations of values detailed in Tables \ref{tab:common-hyperparameters}, \ref{tab:dqn-hyperparameters}, and \ref{tab:PPO_hyperparameters}. This exploration included testing 4 different episode counts, 3 learning rates, 2 random seeds, and 7 batch sizes for common parameters (Table \ref{tab:common-hyperparameters}). For DQN, we explored 4 values for $\gamma$, 4 for initial $\epsilon$, and various other algorithm-specific parameters (Table \ref{tab:dqn-hyperparameters}). Similarly for PPO, we tested 5 values for $\lambda$, 3 for $\epsilon$ in the policy clip, and multiple configurations for network architectures (Table~\ref{tab:PPO_hyperparameters}). This systematic approach resulted in a large number of distinct experiments that covered a wide range of configurations for each algorithm (DQN and PPO) and reward function (symmetric, asymmetric, and hyperbolic). For each scenario, we performed two analyses: one on the hyperparameters and another revealing the best models achieved. Our initial analysis focused on identifying the most crucial hyperparameters and determining which configurations for DQN and PPO yield optimal outcomes. In addition, our goal was to gain insight into the performance of each reward function by individually examining them.

We utilize correlation matrices, a powerful mathematical tool, to analyze the extent to which hyperparameters are correlated with the best score. We only consider correlations with an absolute value greater than 0.15. Any results that appear promising or align with the correlation are then selected.

In the following tables, we present the scores achieved by our DRL models during testing. These scores represent the cumulative rewards obtained by the agent over an episode, indicating how well the model performs in maintaining the desired temperature profile. Higher scores indicate better performance. The Symmetric and Asymmetric reward functions typically yield scores in the hundreds, while the Hyperbolic reward function produces scores between 0 and 1 due to its different scaling. It's important to note that these scores are directly comparable only within the same reward function type.

In a typical production scenario, certain correlations have been observed that hold a significant importance. These correlations are measured and exhibit a noticeable connection between different parameters and the resulting performance. Table \ref{tab:correlation_coefficients} summarizes the results of these findings.

\begin{table}
\centering
\caption{Correlation Coefficients in Normal Production Scenario}
\begin{tabularx}{\textwidth}{|X|c|X|}
\hline
\textbf{Parameter} & \textbf{Correlation Coefficient} & \textbf{Interpretation} \\ \hline
PPO Epochs & 0.4230 & Positive Correlation \\ \hline
PPO Training Interval & 0.3923 & Positive Correlation \\ \hline
PPO Actor Fully Connected 1 Neurons & -0.3262 & Inverse Correlation (Lower value, higher score) \\ \hline
PPO $\varepsilon$ in Policy Clip & 0.3161 & Positive Correlation \\ \hline
PPO c1 in Loss & 0.2653 & Positive Correlation \\ \hline
Use Forge Sensors & -0.1936 & Inverse Correlation (Using forge sensors, lower score) \\ \hline
DQN Transitions Memory & -0.1904 & Inverse Correlation \\ \hline
PPO Critic Fully Connected 1 Neurons & -0.1771 & Inverse Correlation \\ \hline
DQN $\varepsilon$ in $\varepsilon$-greedy & -0.1529 & Inverse Correlation \\ \hline
\end{tabularx}
\label{tab:correlation_coefficients}
\end{table}

These correlations are derived from the analysis of the PPO DQN algorithms within the specified production scenario. The coefficients provide insights into how changes in these parameters could affect the overall performance of the algorithms. Positive coefficients indicate a direct relationship, where an increase in the parameter correlates with an increase in performance. In contrast, negative coefficients suggest an inverse relationship, implying that a decrease in the parameter value could potentially lead to improved performance. The data specifically highlight the nuanced effects of the neural network architecture (such as the number of neurons in fully connected layers) and algorithmic hyperparameters (such as $\varepsilon$ in the policy clip and c1 in loss) on the efficacy of the machine learning models.

Table \ref{tab:PPO_epochs} displays the outcomes achieved using various hyperparameters, namely (PPO) Epochs, (PPO) Training interval, Use forge sensors, and $\varepsilon$ in $\varepsilon$-greedy. This table presents the average scores the PPO algorithm achieves for different numbers of training epochs in the normal production scenario. The scores are shown for three different reward functions: Symmetric, Asymmetric, and Hyperbolic. Higher scores indicate better performance, with the Symmetric and Asymmetric functions typically yielding scores in the hundreds, while the Hyperbolic function produces scores between 0 and 1. The variation in scores across epochs demonstrates the impact of the number of training iterations on model performance. The results indicate that the score increases with an increasing number of training epochs. This could be attributed to the fact that the Actor and Critic DNNs can reduce the agent's loss more efficiently. Additionally, it is worth noting that the average result obtained by the jobs with the hyperbolic reward function is considerably high.

\begin{table}
\centering
\caption{PPO Epochs Analysis in normal production: Test scores achieved with different reward functions. Scores represent cumulative rewards over an episode, with higher values indicating better performance in maintaining the desired temperature profile.}
\begin{tabular}{|c|c|c|c|}
\hline
\textbf{(PPO) Epochs} & \textbf{Symmetric} & \textbf{Asymmetric} & \textbf{Hyperbolic} \\ \hline
5 & 439.711 & 329.045 & 0.129 \\ \hline
10 & 529.712 & 477.993 & 0.216 \\ \hline
15 & 472.257 & 111.429 & 0.324 \\ \hline
20 & 559.902 & 684.969 & 0.522 \\ \hline
25 & 446.595 & 689.504 & 0.227 \\ \hline
30 & 504.743 & 644.979 & 0.643 \\ \hline
\end{tabular}
\label{tab:PPO_epochs}
\end{table}

Table \ref{tab:PPO_intervals} shows the average scores achieved by the PPO algorithm for different training intervals in the normal production scenario. The training interval represents the number of steps taken before retraining the model. Scores are presented for three reward functions: Symmetric, Asymmetric, and Hyperbolic. Higher scores indicate better performance, with Symmetric and Asymmetric functions typically yielding scores in the hundreds, while the Hyperbolic function produces scores between 0 and 1. The variation in scores across training intervals illustrates how the frequency of model updates affects performance for each reward function. The results are presented with respect to the training interval of the agent, specifically, the frequency at which the agent retrains DNN. Within the PPO, this training interval is intrinsically linked to the number of transitions accumulated in the PPO's transition memory. A discernible trend emerges from the data: as the number of steps (and consequently, the number of transitions) increases, there is a corresponding rise in the average score achieved by the agent.

\begin{table}
\centering
\caption{PPO Training Interval Analysis: Test scores for different reward functions in normal production. Scores indicate the model's ability to maintain the desired temperature profile, with higher values representing better performance.}
\begin{tabular}{|c|c|c|c|}
\hline
\textbf{(PPO) Training interval} & \textbf{Symmetric} & \textbf{Asymmetric} & \textbf{Hyperbolic} \\ \hline
10 & 731.871 & 301.635 & 0.171 \\ \hline
25 & 572.918 & 287.197 & 0.486 \\ \hline
50 & 439.64 & 268.613 & 0.366 \\ \hline
100 & 485.41 & 542.401 & 0.123 \\ \hline
200 & 379.149 & 665.818 & 0.525 \\ \hline
500 & 524.616 & 130.409 & 0.193 \\ \hline
1000 & 398.955 & 687.44 & 0.427 \\ \hline
2000 & 498.443 & 684.969 & 0.313 \\ \hline
5000 & 606.455 & 691.889 & 0.151 \\ \hline
\end{tabular}
\label{tab:PPO_intervals}
\end{table}

For the scenario after warmholding, the hyperparameters with higher correlation coefficients are as in Table \ref{tab:hyperparameter_afterwarmholding}. Finally, the results obtained with the following hyperparameters: (PPO) Actor F.C. 1 neurons, No noise before zone 3, (PPO) c1 in loss. The others are not presented, as they do not significantly contribute to the performance. As in the case of Tables 6 and 7, Table \ref{tab:actor_fc1_neurons} shows the average scores obtained with the various values of neurons in the first fully connected layer of the Actor DNN. The correlation coefficient matches the results, except for the average score using 512 neurons with the hyperbolic reward function.

\begin{table}
\centering
\caption{Hyperparameters and settings for after warmholding scenario}
\label{tab:hyperparameter_afterwarmholding}
\begin{tabular}{|l|l|}
\hline
\textbf{Parameter} & \textbf{Value} \\ \hline
(DQN) \( \gamma \) & -0.3845 \\ \hline
(PPO) Actor fully connected 1 neurons & 0.3545 \\ \hline
(PPO) Actor fully connected 2 neurons & 0.3292 \\ \hline
No noise before zone 3 & -0.3153 \\ \hline
(PPO) \( c_1 \) in loss & 0.1786 \\ \hline
\end{tabular}
\end{table}

\begin{table}
\centering
\caption{PPO Actor Network Analysis: Test scores for different reward functions in normal production after warmholding. Scores reflect the model's performance in maintaining the desired temperature profile after transitioning from warmholding, with higher values indicating better performance.}
\label{tab:actor_fc1_neurons}
\begin{tabular}{|c|c|c|c|}
\hline
\textbf{(PPO) Actor F.C. 1 neurons} & \textbf{Symmetric} & \textbf{Asymmetric} & \textbf{Hyperbolic} \\ \hline
128 & 97.422 & 161.654 & 0.165 \\ \hline
256 & 183.306 & 172.646 & 0.254 \\ \hline
512 & 196.603 & 181.276 & 0.098 \\ \hline
\end{tabular}
\end{table}

\subsubsection{Best Model}

We conducted a series of experiments to identify the optimal model by fine-tuning hyperparameters for each algorithm. In Table \ref{tab:BestModelNormalPRoduction}, we present the findings of these experiments in relation to the normal production scenario. On the basis of these results, we have determined that the model generated with PPO and the symmetric reward function is the most effective. This decision was not based on its performance during training (as DQN with a symmetric reward function achieved the best score), but rather on its consistently high mean score during testing with both the symmetric and hyperbolic functions, indicating that it closely aligns with the desired temperature profile.

\begin{table}
\centering
\caption{Best models obtained in the proper normal production scenario}
\begin{tabular}{|c|c|c|c|c|}
\hline
\textbf{Algo.} & \textbf{Rew. fun.} & \textbf{Best score} & \textbf{Mean score (same fun.)} & \textbf{Mean score (hyperb. fun.)} \\ \hline
DQN & Sym. & 796.469 & 704.804 & 0.510 \\ \hline
DQN & Asym. & 689.968 & 709.283 & 0.475 \\ \hline
DQN & Hyperb. & 0.621 & 0.547 & 0.443 \\ \hline
PPO & Sym. & 792.307 & 725.609 & 0.542 \\ \hline
PPO & Asym. & 727.665 & 756.205 & 0.526 \\ \hline
PPO & Hyperb. & 0.643 & 0.538 & 0.469 \\ \hline
\end{tabular}
\label{tab:BestModelNormalPRoduction}
\end{table}

\begin{figure}
    \centering
    \includegraphics[width=0.75\linewidth]{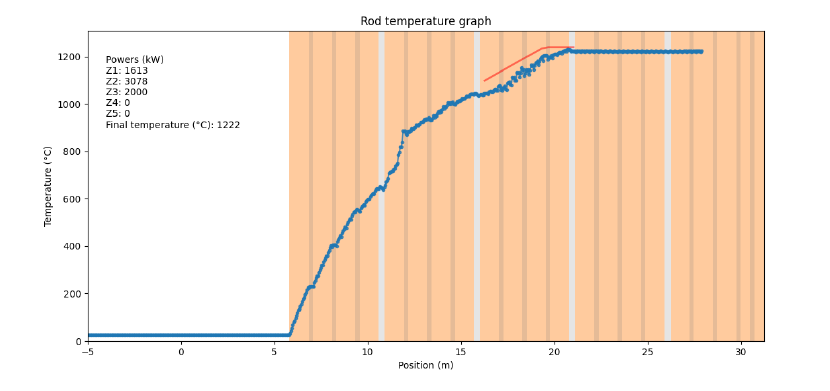}
    \caption{Running the model in proper normal production}
    \label{fig:RunningModeNormalProduction}
\end{figure}

Table \ref{tab:hyperparameters_best_model} shows the hyperparameters with which the best model was trained. To know how the best model performs in normal production, Figure \ref{fig:RunningModeNormalProduction} shows the temperature profile and the running results of the best model in normal production.

\begin{table}
\centering
\caption{Hyperparameters of the best model in proper normal production.}
\label{tab:hyperparameters_best_model}
\begin{tabular}{|l|l|l|l|}
\hline
\textbf{Hyperparameter} & \textbf{Value} & \textbf{Hyperparameter} & \textbf{Value} \\ \hline
Episodes & 100 & \( c_1 \) in loss & 1 \\ \hline
Learning rate & 0.001 & \( \varepsilon \) in policy clip & 0.2 \\ \hline
Seed & 19 & Epochs & 20 \\ \hline
Batch size & 4096 & Training interval & 100 \\ \hline
Normalization & Enabled & Actor F.C. 1 neurons & 256 \\ \hline
Z1 and Z2 powers & Without noise & Actor F.C. 2 neurons & 512 \\ \hline
Sensors & Virtual & Critic F.C. 1 neurons & 256 \\ \hline
\( \lambda \) for GAE & 1 & Critic F.C. 2 neurons & 256 \\ \hline
\end{tabular}
\end{table}

As for the best model in the normal production after warmholding scenario, Table \ref{tab:best_models_Afterwarmholding} show the comparison of the score for both DQN and PPO algorithms. Although none of the models presented in this table exhibit clear superiority over the others, it is worth noting that the DQN model with hyperbolic function performed well and shows promising potential. Therefore, it has been selected as the optimal model. For reference, the hyperparameters used during its training are listed in Table \ref{tab:hyperparameters_best_model_AfterWarmholding}.

\begin{table}
\centering
\caption{Best models obtained in the normal production after warmholding scenario.}
\label{tab:best_models_Afterwarmholding}
\begin{tabular}{|c|c|c|c|c|}
\hline
\textbf{Algo.} & \textbf{Rew. fun.} & \textbf{Best score} & \textbf{Mean score (same fun.)} & \textbf{Mean score (hyperb. fun.)} \\ \hline
DQN & Sym. & 229.112 & 227.932 & 0.281 \\ \hline
DQN & Asym. & 208.230 & 170.357 & 0.157 \\ \hline
DQN & \textit{Hyperb.} & 0.307 & 0.274 & 0.298 \\ \hline
PPO & Sym. & 228.791 & 231.479 & 0.293 \\ \hline
PPO & Asym. & 209.637 & 204.991 & 0.279 \\ \hline
PPO & \textit{Hyperb.} & 0.301 & 0.301 & 0.270 \\ \hline
\end{tabular}
\end{table}

\begin{table}
\centering
\caption{Hyperparameters of the best model in normal production after warmholding.}
\label{tab:hyperparameters_best_model_AfterWarmholding}
\begin{tabular}{|l|l|l|l|}
\hline
\textbf{Hyperparameter} & \textbf{Value} & \textbf{Hyperparameter} & \textbf{Value} \\ \hline
Episodes & 300 & F.C. 2 neurons & 256 \\ \hline
Learning rate & 0.00001 & Target net. update interval & 100000 \\ \hline
Seed & 19 & Transitions memory & 100000 \\ \hline
Batch size & 1024 & \( \epsilon \) in \( \epsilon \)-greedy & 0.7 \\ \hline
Normalization & Disabled & \( \epsilon \)'s lower bound & 0.01 \\ \hline
Z1 and Z2 powers & With noise & \( \epsilon \)'s decrease step & 0.05 \\ \hline
Sensors & Virtual & \( \gamma \) & 0.9 \\ \hline
F.C. 1 neurons & 512 & - & - \\ \hline
\end{tabular}
\end{table}

\begin{figure}
    \centering
    \includegraphics[width=0.75\linewidth]{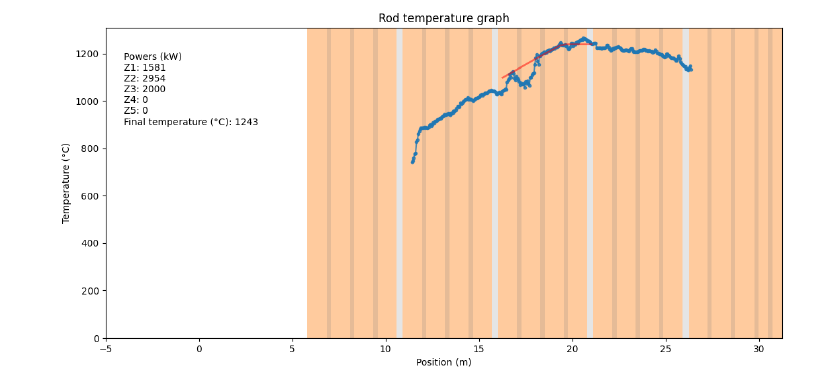}
    \caption{Running the model in normal production after warmholding}
    \label{fig:BestModelGraphAfterWarmHolding}
\end{figure}

The empirical data presented in Figure \ref{fig:BestModelGraphAfterWarmHolding} provides a snapshot of the model's performance during testing. The model demonstrated a higher degree of ability to fit the desired temperature profile.

\section{\textcolor{blue}{Impact, Generalizability, and Lessons Learned}}\label{ImpactOnSustain}

\textcolor{blue}{Our integrated approach—combining DT modeling, EDMA for real-time processing, and DRL for intelligent control, has shown significant potential to improve steel production processes. While each component contributes unique value, it is their combined implementation that enables the improvements discussed below.}

Our DT-based approach, which uses DRL, has \ak{the potential for} significant improved steel production processes' control. Although direct measurements of sustainability, efficiency, and cost-effectiveness are not provided, we can infer these improvements from our experimental results. \textcolor{blue}{ The measurement of waste reduction in industrial settings like steel production faces several challenges that prevented direct quantification in this study. These include production priorities that often limit opportunities for experimental control groups, operational constraints that restrict prolonged data collection periods, resource limitations affecting sensor deployment, and the complex nature of waste, which can manifest as energy inefficiency, material scrap, or quality defects. In addition, proprietary concerns in industrial partnerships often limit the disclosure of certain performance metrics. Despite these constraints, our experimental results provide strong indirect evidence of \ak{the potential} improvements through the temperature profile optimizations shown in our testing.}

\subsection{Efficiency and Quality Improvements}

The DRL-powered control system has shown substantial improvements in maintaining optimal temperature profiles during steel production \ak{on the digital twin}. In the normal production scenario, our best model achieved a high degree of alignment with the desired temperature profile, as evidenced in Figure \ref{fig:RunningModeNormalProduction}. This precise temperature control likely leads to improved product quality and reduced energy waste from overheating \ak{once deployed on the actual system}. Furthermore, in the challenging scenario of normal production after warmholding, Figure \ref{fig:BestModelGraphAfterWarmHolding} demonstrates our model's ability to effectively manage temperature profiles even when starting from non-uniform temperature distributions. This adaptability suggests improved efficiency \ak{when deployed in the factory} in managing production variations and potentially reduced waste from suboptimal heating patterns.

\subsection{Sustainability and Cost-Effectiveness Implications}

The precise temperature control demonstrated in both scenarios indicates more efficient energy use in heating. Our system \ak{has the potential to} reduce unnecessary energy consumption and associated costs by maintaining optimal temperatures without overheating. The improved control, particularly in challenging scenarios like post-warmholding, suggests a potential reduction in material waste due to temperature-related quality issues. Furthermore, our DRL approach demonstrated faster convergence to optimal temperature profiles, which reduces overall processing time and associated energy costs. The precise control also suggests a potential increase in the percentage of products that meet quality standards, thereby improving the yield and reducing the costs associated with rework or scrap.

\subsection{Use Case Scenario: Warmholding Optimization at Bharat Forge Kilsta}\label{usecase_replication}

We implemented our DRL-powered digital twin approach at Bharat Forge Kilsta AB to address the challenge of uneven temperature distributions in steel rods after warmholding, which may lead to a "zebra pattern'' problem if not addressed properly. Using data from their furnace sensors, we initialized our model. We demonstrated a significant improvement in the management of temperature profiles during the transition from warmholding to normal production \ak{in the digital twin}, as shown in Figure 14. This real-world application potentially reduces time, energy consumption, and material waste \ak{once deployed} in their production process.

Due to the sensitive nature of industrial processes, we cannot share the full implementation details or source code. However, researchers can reproduce a similar approach by initializing a DT with warmholding temperature patterns (Section \ref{section4.6}), applying the DRL model framework (Sections \ref{DRLpowedControl}). While exact replication may not be possible, these guidelines provide a foundation for adapting our methodology to similar steel production scenarios, balancing reproducibility with the necessary industrial confidentiality.

\subsection{\textcolor{blue}{Generalizability and Lessons Learned}}

\textcolor{blue}{While our implementation is tailored to steel production at Bharat Forge Kilsta, several key lessons and generalizable approaches emerged from this work that can benefit other industries and applications:}

\begin{enumerate}
    \item \textcolor{blue}{Architectural Patterns for Industrial ML: The separation of concerns in our MLOps-driven architecture (data input layer, event-driven microservices, and model management) provides a blueprint that can be adapted across various industrial control systems beyond steel production. This pattern is particularly valuable in environments with strict latency and security requirements and legacy equipment integration needs.}
   
    \item \textcolor{blue}{DT Integration Strategy: Our approach to building a DT that supports both simulation and real-time control offers a framework that can be applied to other continuous manufacturing processes, such as chemical processing, paper manufacturing, or glass production. The design choices for simulator managers (Section 4.6) represent a generalizable pattern for building adaptable DT.}
   
    \item \textcolor{blue}{MLOps Workflow in Industrial Settings: The deployment and model management approach described in Section \ref{Deployment} provides practical guidance for managing ML models in industrial settings with limited connectivity, emphasizing model wrapping techniques that can be applied regardless of the specific industry.}
    
    \item \textcolor{blue}{Reward Function Design Insights: The comparative analysis of reward functions (Section \ref{section9.2}) offers valuable insights for other RL applications in industrial control, showing how different reward formulations can significantly impact control quality and adaptation capabilities.}
    
    \item \textcolor{blue}{Edge Computing Requirements: Our work demonstrates the practical necessity of edge computing for industrial ML applications with strict latency requirements, providing a reference architecture that balances computational needs with deployment constraints common across manufacturing environments.}
    
\end{enumerate}

\section{Conclusion and Future Work}\label{Conclusion}

\textcolor{blue}{In this paper, we have presented an innovative approach that integrates DT technology, Event-Driven Microservices Architecture, and DRL to revolutionize steel production. Each component played a vital role: the DT provided a virtual representation of the physical process, the EDMA enabled real-time data processing and model deployment, and DRL algorithms optimized control decisions within this framework. In addition, }we have presented an innovative approach that leverages the power of DRL and MLOps to revolutionize steel production. Our primary focus has been automating the normal production mode, recognizing that various algorithms may be required for different production modes. Instead of going into algorithmic complications, we prioritized the architecture and infrastructure that form the backbone of our solution. The application of DRL to control the electrical power consumed during the heating process of an induction furnace has yielded remarkable results. Our system has \ak{has the potential to}
substantially reduce waste, improve environmental sustainability, and improve energy efficiency by continuously adapting to dynamic operational conditions in real-time. These results align with the growing demand for sustainable and environmentally friendly manufacturing practices. Moreover, our work serves as a bridge between two distinct domains: machine learning and industrial operations. We have demonstrated that MLOps can play a key role in orchestrating the transformation of traditional industrial processes into smart, autonomous systems. This integration allows for effectively deploying and managing machine learning models in production environments.

While our research has made significant strides in reshaping the steel production landscape, it is important to recognize that there are several avenues for future work and exploration. Future research should focus on developing and integrating algorithms for different production modes beyond normal operation, such as optimizing start-up procedures, handling emergency shutdowns, and managing maintenance cycles. Additionally, exploring multi-objective optimization techniques could enable simultaneous consideration of factors like product quality, equipment lifespan, and production schedules alongside our current focus on energy efficiency and waste reduction. The potential of transfer learning in industrial settings is another promising area, which could allow the application of our DRL models to  
different types of furnaces or even other steel production processes with minimal retraining. Developing methods to make the decisions of our DRL models more interpretable to human operators through explainable AI techniques could enhance trust and facilitate wider adoption in industrial settings. In terms of advancing our MLOps practices specifically, future work should focus on enhancing reproducibility through comprehensive version control systems for code, data, and models. We also aim to explore the integration of Continuous Integration and Continuous Deployment (CI/CD) practices to automate model building, testing, and deployment processes, streamlining our workflow and ensuring consistent quality in model updates. Another critical area for future investigation is the development of robust systems for monitoring and tracking the performance of our models in production settings, including real-time performance metrics, drift detection, and automated alerts for model degradation.

Further research is needed to test the scalability of our approach to \ak{the real production facility} larger production facilities and its generalizability to other heavy industries beyond steel production. Additionally, we plan to investigate techniques for automated model retraining based on performance metrics and new data, ensuring our models remain effective over time. Future work should also include more rigorous statistical analysis of the correlations between hyperparameters and algorithm performance, potentially employing advanced statistical methods suitable for multi-dimensional hyperparameter spaces. Finally, extended studies are required to assess the long-term performance of the system and its ability to adapt to gradual changes in equipment efficiency or production requirements over time.

Heavy industry, including steel production, remains an underexplored domain in MLOps. More studies are needed that not only explore the theoretical aspects, but also present concrete results and lessons learned from the implementation of customized MLOps solutions. Our work underscores the transformative potential of MLOps and DRL in industrial settings, emphasizing the importance of focusing on infrastructure and architecture to show the way for smarter and more efficient production processes. The future directions outlined above represent exciting opportunities to build on this foundation and further advance the intelligent manufacturing field, paving the way for more sustainable, efficient, and intelligent industrial processes. To support the replicability of our work, we have provided a detailed use case scenario in Section \ref{usecase_replication}, which offers guidelines for reproducing our approach in similar industrial settings. Although full replication may be challenging due to the sensitive nature of industrial data, we believe that this use case, combined with the detailed descriptions throughout the paper, provides a solid foundation for adapting our methods to other contexts in heavy industry.

\section*{Acknowledgement}

This work was partially funded by Vinnova through the SmartForge project. Additional funding was provided by the Knowledge Foundation of Sweden (KKS) through the Synergy Project AIDA - A Holistic AI-driven Networking and Processing Framework for Industrial IoT (Rek:20200067) and by the Bavarian State Ministry of Education and Culture, Science and Art through the HighTech Agenda (HTA).

\bibliographystyle{elsarticle-num} 
\bibliography{References}

\end{document}